\pdfoutput=1
\documentclass[11pt]{article}

\usepackage[preprint]{acl}

\usepackage{times}
\usepackage{latexsym}

\usepackage[T1]{fontenc}

\usepackage[utf8]{inputenc}

\usepackage{microtype}

\usepackage{inconsolata}

\usepackage{graphicx}
\usepackage{amsmath}
\usepackage{amssymb}
\usepackage{booktabs}
\usepackage{multirow}
\usepackage{xcolor}
\usepackage{tikz}
\usetikzlibrary{positioning,arrows.meta}
\usepackage{algorithm}
\usepackage{algpseudocode}
\usepackage{cleveref}
\usepackage{xspace}
\hypersetup{
  pdftitle={SocialGym: Evaluating LLM Social Intelligence through Verifiable Multi-Agent Games},
  pdfauthor={Keyu He, Xuhui Zhou, Maarten Sap}
}

\newcommand{\socialgym}{\textsc{\mbox{Social}Gym}\xspace}

\crefformat{subsection}{§#2#1#3}
\Crefformat{subsection}{§#2#1#3}
\crefname{appendix}{Appendix}{Appendices}
\Crefname{appendix}{Appendix}{Appendices}

\title{\socialgym: Evaluating LLM Social Intelligence through Verifiable Multi-Agent Games}

\author{
  \textbf{Keyu He} \quad \textbf{Xuhui Zhou} \quad \textbf{Maarten Sap} \\
  Carnegie Mellon University \\
  \texttt{keyuhe@cmu.edu}
}

\begin{document}
\maketitle
\begin{abstract}
Large Language Model (LLM) agents are increasingly deployed in multi-agent social settings where they must cooperate, negotiate, and adapt to other agents.
Measuring these social skills is hard because most social interactions offer no objective ground truth. Evaluations often fall back on LLM judges, which are costly, subjective, and noisy.
We introduce \textbf{\socialgym}, an environment of 22 multi-agent social games, from matrix games to hidden-role deduction, with verifiable outcomes determined by game rules.
We benchmark ten LLMs across all games and roles using an Elo tournament.
Our results show that no model is best in every game, and rankings shift across game families and sides.
Secret keeping correlates more strongly with theory-of-mind and privacy benchmarks than with general knowledge benchmarks.
Half of the models give away their side's secret in Chameleon and Spyfall, and the smallest models are not reliably better than chance at voting for hidden opponents in most hidden-role games.
Finally, we examine whether models can improve their play by reflecting on self-play and writing strategy playbooks, finding mixed results across models and games.
\socialgym provides verifiable measures for diagnosing LLM social capabilities across settings and roles.
\end{abstract}

\section{Introduction}
\label{sec:intro}

LLM-based agents are increasingly deployed and studied as social agents in multi-party interactions with humans or other agents: serving as group mediators and companions for humans~\citep{tessler2024ai,defreitas2026companions}, collaborating on shared tasks through inter-agent dialogue~\citep{li2023camel,wu2023autogenenablingnextgenllm}, role-playing as autonomous inhabitants of social sandboxes~\citep{park2023generative,piao2025agentsociety}, and acting as persuaders or negotiators against other models~\citep{bianchi2024well}.
These settings require social interaction capabilities beyond single-turn question answering.
Agents must handle information asymmetry \citep{zhou2024realLifeFantasy,ys2026sotopiaTom}, deception-utility tradeoffs \citep{su2025ailiedar}, and negotiations \citep{cohen2026imperfectlyCooperative}, all while sustaining coherent role-play \citep{shao2023character, wang-etal-2024-rolellm}.

Existing evaluations of LLM social reasoning pose two measurement challenges.
First, many evaluations rely on static benchmarks such as theory-of-mind questionnaires~\citep{kim2023fantom}, which produce reproducible scores but cannot test sustained multi-turn behavior.
Second, though more recent open-ended interactive evaluations have been developed~\citep[e.g.,][]{zhou2024sotopiainteractiveevaluationsocial}, they assess multi-turn interaction skills with LLM-as-a-judge scoring, which suffers from position, verbosity, and self-enhancement biases~\citep{zheng2023judging}, and is also relatively variable and subjective \citep{zhou2024sotopiainteractiveevaluationsocial}.

In this work, we turn to social games as a testbed for social skills. 
Such games address both evaluation challenges and thus pose several advantages.
First, game rules deterministically decide every outcome without relying on LLM judges, which opens the door for verifiable outcomes that could ultimately serve as rewards for reinforcement learning~\citep{guo2025deepseek, lambert2024tulu3, liu2026spiralselfplayzerosumgames}.
Second, games test complex social abilities by requiring agents to infer what others know, conceal private information, and act on that information over many turns~\citep{guertler2025textarena,wang2026mindgames}.
Finally, games vary in player count, communication, and access to information, letting us test how agents handle information asymmetry~\citep{zhou2024realLifeFantasy}.

\begin{figure*}[t]
\centering
\includegraphics[width=\textwidth]{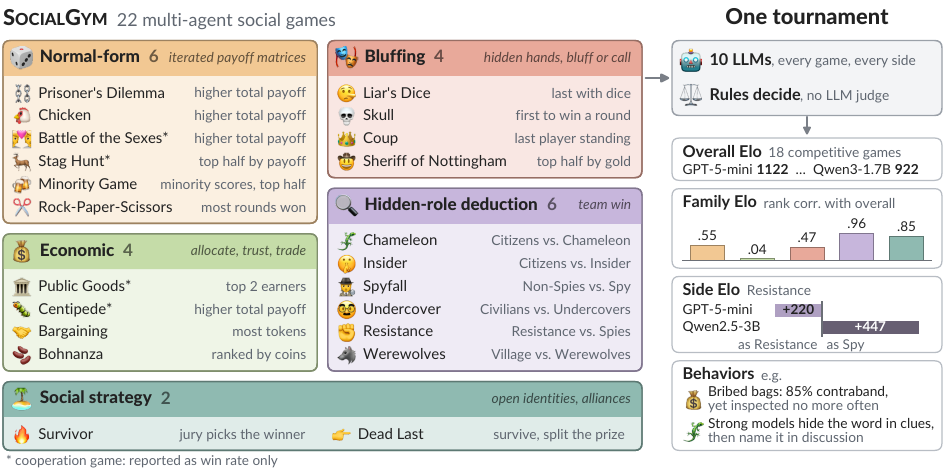}
\caption{Overview of \textbf{\socialgym}: 22 multi-agent social games in five families, each scored by its rules.}
\label{fig:fig1}
\end{figure*}

We introduce \textbf{\socialgym} (\Cref{sec:benchmark}), an environment of 22 multi-agent games in five families that span diverse social scenarios.
The suite includes repeated strategic choice with little or no communication, resource allocation under competing interests, bluffing about hidden state, hidden-role deduction through dialogue, and coalition formation with public identities.
One engine runs all games and gives every agent a role-specific goal with public, team, or private observations.
\socialgym builds on---yet substantially differs from---existing game-based evaluations, which typically only examine one social setting with their own models and protocol~\citep{xu2023exploring, lan2024avalon, akata2025playing}, or consist mostly of single- and two-player games, with few settings of hidden allegiances or shifting coalitions~\citep{duan2024gtbench, guertler2025textarena}.
\socialgym puts the same agents through 22 games, on every side of each game, under one protocol, to test generalizable social skills in multiple types of social interactions.

In our experiments, we benchmark ten models on every game and every role under a common tournament protocol, which yields ratings per game, per family, and per side for hidden-role games (\Cref{fig:fig1}).
Our results show that no model is best across all games, and rankings vary across game families and roles (\Cref{sec:results:separation}), suggesting that evaluating social intelligence requires examining many kinds of games at once.
Secret-keeping performance in our games correlates more strongly with FANToM~\citep{kim2023fantom} and PrivacyLens~\citep{shao2024privacylens} than with MMLU-Pro and GPQA (\Cref{sec:results:reliability}).
These correlations support using secret keeping during play as a behavioral measure of social reasoning.
Our experiments also reveal insights related to model scale; for example, stronger models playing Chameleon keep the secret word out of their clues but often name it during discussion, and the smallest models are not reliably better than chance at identifying hidden opponents in most hidden-role games (e.g., Werewolf, \Cref{sec:results:secrets}).

Finally, we examine whether models can improve their play by reflecting on self-play and writing strategy playbooks, and find mixed results across models and games (\Cref{sec:experiments}).
This shows that prompting alone does not reliably improve social play, and that more in-depth approaches are likely needed.
Our work highlights the complexity of social interaction and the promise of social games as verifiable measures for evaluating and training more socially skilled LLMs.

\section{Related Work} 
\label{sec:related}

Social intelligence combines social knowledge with strategies for applying it~\citep{kihlstrom2000social}.
Existing LLM evaluations take three main approaches.
First, static benchmarks such as FANToM~\citep{kim2023fantom}, ToMi~\citep{le2019tomi}, and BigToM~\citep{gandhi2023understanding} test social reasoning through fixed questions but do not evaluate sustained interaction.
Second, interactive evaluations such as SOTOPIA~\citep{zhou2024sotopiainteractiveevaluationsocial, wang2024sotopia} assess multi-turn social behavior using human ratings or LLM judges, whose scores are subject to documented biases~\citep{zheng2023judging}.
STSS~\citep{wang2024stss} also evaluates social interaction, but scores an agent by task completion in a sandbox populated by LLM-driven characters.

Game-based evaluations, the third approach, provide verifiable outcomes computed from game rules.
Studies of Diplomacy~\citep{meta2022human}, Werewolf~\citep{xu2023exploring}, Avalon~\citep{lan2024avalon, light2023avalonbench}, and repeated matrix games~\citep{akata2025playing} examine individual social settings with different models and evaluation protocols.
Game collections provide shared environments across many games~\citep{guertler2025textarena, duan2024gtbench, wu2024smartplay, chalamalasetti2023clembench}, but most of their games are single- or two-player.
MINDGAMES targets social and strategic reasoning across four environments~\citep{wang2026mindgames}.
In contrast, to compare models across kinds of social interaction, \socialgym organizes 22 games along information and communication axes, including ten multi-party dialogue games.
It rates every model on both sides of each hidden-role game.
We provide a detailed comparison with previous game-based setups in \Cref{tab:compare}.

\begin{table}[t]
\centering
\footnotesize
\resizebox{\linewidth}{!}{
\begin{tabular}{@{}l@{\hskip 4pt}c@{\hskip 4pt}c@{\hskip 4pt}c@{\hskip 4pt}l@{}}
\toprule
Benchmark & \shortstack{Multi-\\party} & \shortstack{Multi-party\\dialogue} & \shortstack{Hidden-\\role} & \shortstack[l]{Per-role\\evaluation} \\
\midrule
\multicolumn{5}{@{}l}{\emph{Single game}} \\
Werewolf Arena~\citep{bailis2024werewolf} & 1 & 1 & 1 & both, mostly in-family \\
AvalonBench~\citep{light2023avalonbench} & 1 & 1 & 1 & both, vs.\ bots \\
\midrule
\multicolumn{5}{@{}l}{\emph{Multiple games}} \\
GAMA-Bench~\citep{huang2025gamabench} & 8 & 0 & 0 & $\times$ \\
MAgIC~\citep{xu2024magic} & 5 & 2 & 2 & both, vs.\ GPT-4 \\
LLMArena~\citep{chen2024llmarena} & 1 & 1 & 1 & one, vs.\ GPT-3.5 \\
SPIN-Bench~\citep{yao2025spinbench} & 2 & 1 & 0 & $\times$ \\
TextArena~\citep{guertler2025textarena} & 11 & 5 & 1 & $\times$ \\
MINDGAMES~\citep{wang2026mindgames} & 3 & 2 & 1 & $\times$ \\
\midrule
\socialgym (ours) & 16 & 10 & 6 & both, vs.\ models across \\
 & & & & families and sizes \\
\bottomrule
\end{tabular}
}
\caption{\socialgym and prior game-based evaluations of LLMs.
Multi-party: games with more than two players.
Multi-party dialogue: those in which players exchange messages.
Hidden-role: social deduction games with secret teams.
Per-role evaluation: which sides of a team game the model plays, and against whom.}
\label{tab:compare}
\end{table}

\section{\texorpdfstring{\socialgym}{SocialGym}: Games and Evaluation}
\label{sec:benchmark}

We design \socialgym around two principles: \textbf{(i)} every game must have an \emph{algorithmically verifiable} outcome (win/loss, score, survival), providing an unambiguous and verifiable evaluation signal; 
and \textbf{(ii)} games must cover different social challenges, including cooperation, coordination, negotiation, and deception, to expose distinct failure modes.

\subsection{Game Design and Coverage}
\label{subsec:game_taxonomy}

\begin{figure*}[t]
\centering
\includegraphics[width=\textwidth]{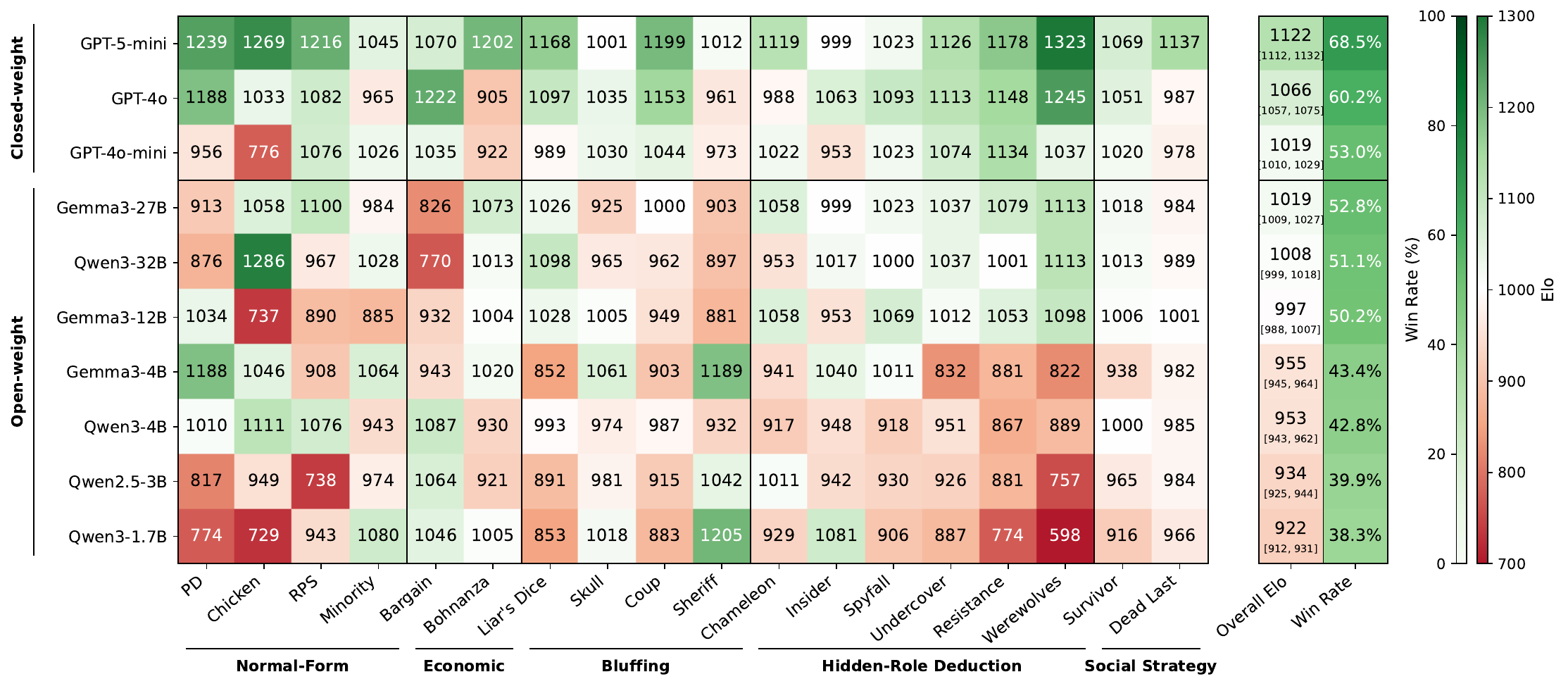}
\caption{Per-game Elo on the 18 competitive games (left) and overall Elo with 95\% bootstrap CIs and win rate (right).
Columns are grouped by family and rows sorted by overall Elo.
A tie counts as half a win.
Each model plays 54 episodes per game (56 in Coup and Skull, 90 in Dead Last), 1{,}012 in total.}
\label{fig:per_game_heatmap}
\end{figure*}

We group our 22 games into five families by the social challenge each poses.
Two axes describe how the games differ, \emph{information structure} (complete information / hidden state / hidden roles) and \emph{communication mode} (none / structured / free-form; \Cref{tab:taxonomy}).
Each family probes a different facet of social intelligence, grounded in strategic primitives \citep{axelrod1981evolution,schelling1980strategy}, prosocial behavior under collective-action problems \citep{ostrom1990governing,fehr2000cooperation}, sequential trust and bargaining \citep{rosenthal1981games,rubinstein1982perfect}, deception production and detection \citep{depaulo2003cues,wimmer1983beliefs}, theory of mind \citep{baroncohen1985autistic,PERNER1985437,byrne1988machiavellian}, and coalition and reputation tracking \citep{dunbar1998social}, including persuasion \citep{cialdini1984influence}.

\paragraph{Normal-Form Games (6).}
Iterated matrix games with complete information and no communication, except for Battle of the Sexes, which we play with a pre-play discussion, a variant studied by \citet{cooper1989communication}. 
\emph{Games:} Prisoner's Dilemma, Chicken, Battle of the Sexes, Stag Hunt, Minority Game, Rock-Paper-Scissors. 
Prisoner's Dilemma and Battle of the Sexes follow \citeauthor{akata2025playing}'s study of LLMs on repeated $2{\times}2$ matrix games~\citep{akata2025playing}, and our Stag Hunt has four players.
We extend the suite with Minority Game and Rock-Paper-Scissors and embed all six in a unified leaderboard.

\paragraph{Economic Games (4).}
Multi-round resource-allocation games requiring strategic reasoning and (optionally) negotiation. 
\emph{Games:} Public Goods Game (free-riding vs.\ collective action), Centipede (sequential trust under growing stakes), Bargaining (structured negotiation over a divisible payoff), and \href{https://boardgamegeek.com/boardgame/11/bohnanza}{Bohnanza} (free-form multilateral card trading under a fixed hand order, with public fields and private hands).

\paragraph{Bluffing Games (4).}
Hidden-state games (no fixed factions) where agents must misrepresent or correctly infer private information. 
\emph{Games:} \href{https://boardgamegeek.com/boardgame/45/liars-dice}{Liar's Dice} (probabilistic reasoning with hidden dice), \href{https://boardgamegeek.com/boardgame/92415/skull}{Skull} (placement bluffing with no communication), \href{https://boardgamegeek.com/boardgame/131357/coup}{Coup} (structured action-claim bluffing), \href{https://boardgamegeek.com/boardgame/157969/sheriff-of-nottingham}{Sheriff of Nottingham} (free-form negotiation under inspection).

\paragraph{Hidden-Role Deduction (6).}
Games with hidden role assignments where agents must identify allies and enemies through dialogue. 
\emph{Games:} \href{https://boardgamegeek.com/boardgame/227072/the-chameleon}{Chameleon}, \href{https://boardgamegeek.com/boardgame/206051/insider}{Insider}, \href{https://boardgamegeek.com/boardgame/166384/spyfall}{Spyfall}, \href{https://yanstarstudio.com/undercover-how-to-play}{Undercover}, \href{https://boardgamegeek.com/boardgame/41114/the-resistance}{Resistance}, \href{https://boardgamegeek.com/boardgame/25821/the-werewolves-of-millers-hollow}{Werewolves}.
Prior work studies these games one at a time, such as Werewolf~\citep{xu2023exploring, bailis2024werewolf} and Avalon~\citep{lan2024avalon, light2023avalonbench}, a variant of The Resistance.
We evaluate all six under one protocol.

\paragraph{Social Strategy (2).}
Complete-information games where outcomes depend on alliance formation, persuasion, and reputation, with no hidden information. 
\emph{Games:} \href{https://en.wikipedia.org/wiki/Survivor_(American_TV_series)}{Survivor} (jury-voted finals), \href{https://boardgamegeek.com/boardgame/184522/dead-last}{Dead Last} (elimination + final-round split).

\paragraph{Player formats.}
Across families, the games come in three player formats.
In \emph{two-player games}, such as Prisoner's Dilemma, two players face each other.
In \emph{two-team games}, the six hidden-role games, four to six players split into two teams.
In \emph{free-for-all games}, such as Coup and Survivor, three to six players each play for themselves.
\Cref{tab:taxonomy} gives the format of every game.

\subsection{System Architecture}
\label{subsec:sys_arch}
\socialgym also includes an engine that runs all 22 games.
It extends the SOTOPIA environment loop~\citep{zhou2024sotopiainteractiveevaluationsocial} to $N$ agents, with a finite-state machine for game phases such as Night and Day in Werewolves.
Each player role can be assigned a different LLM through LiteLLM, which provides a shared interface for API-hosted and locally served models.
Every message is public, visible to one team, or private to one agent, so agents must reason about what others can see.
The rules compute every outcome, and all games share one interface (\Cref{sec:appendix:engine}).
The code is public (\Cref{sec:appendix:code}).

\subsection{Evaluation Protocol}
\label{subsec:elo}
In multi-agent LLM simulations, the models behind the other agents shape how an interaction unfolds~\citep{roytburg2026disentangling}, so a model's score depends on whom it plays.
We therefore use a round-robin tournament to evaluate each model against a balanced set of opponents.
Every model plays every game in every role against agents powered by the other models, with no self-play.
The schedule depends on the player format (\Cref{subsec:game_taxonomy}).
In two-player games, each pair of models meets equally often.
In two-team games, one model fills every seat of a team, and each pair plays equally often with each model on each team.
In free-for-all games, every seat goes to a different model, and all pairs share about the same number of tables.
The schedule balances opponent pairings and role assignments, and seat order is shuffled per episode.

The rules give every agent a score at the end of each episode.
In two-player and free-for-all games, every cross-model pair of agents yields one outcome, and the higher score wins.
In two-team games, only cross-team pairs count, using each team's shared score.
\socialgym fits Elo-scale ratings~\citep{elo1978rating} to these outcomes with a regularized Bradley--Terry model~\citep{bradley1952rank}, as in Chatbot Arena (\citealp{chiang2024chatbot}; \Cref{sec:appendix:elo_details}).
It reports ratings per game, per family, and over the 18 competitive games.
In hidden-role games, it also rates each side separately, \textbf{Elo-Main} for the majority (Villager, Civilian, Non-Spy) and \textbf{Elo-Alt} for the minority (Werewolf, Spy, Insider, Chameleon, Undercover).
For the four cooperation and coordination games, \socialgym reports pairwise win rates instead of Elo (\Cref{tab:coop_winrate}).

\section{Benchmarking Social Behavior}
\label{sec:results}
We benchmark three closed-weight models (GPT-5-mini, GPT-4o, GPT-4o-mini) and seven open-weight models ranging from 1.7B to 32B parameters (Gemma3-27B, 12B, and 4B; Qwen3-32B, Qwen3-4B, Qwen2.5-3B, and Qwen3-1.7B; \Cref{sec:appendix:models}).
In our tournament, each model plays 54 episodes of each game, or 1{,}228 in total, six per opponent in two-player and two-team games.\footnote{Coup and Skull have 56 episodes per model and Dead Last 90, with less even pair counts.}
The tournament has 4{,}588 episodes, and the rankings are stable at this size (\Cref{sec:appendix:stability}).
Each model plays 1{,}012 episodes of the 18 competitive games, which give about 3{,}700 pairwise outcomes per model.
Every rating has a 95\% bootstrap confidence interval (\Cref{sec:appendix:elo_details}).
API costs are in \Cref{sec:appendix:cost}.

We compare the models across game families and roles (\Cref{sec:results:separation}), relate their ratings to outside benchmarks and an LLM judge (\Cref{sec:results:reliability}), and examine how they keep secrets, vote, respond to bribes, and trade (\Cref{sec:results:secrets}).

\subsection{Performance across social settings and roles}
\label{subsec:leaderboard}
\label{sec:results:separation}

\paragraph{Overall.}
GPT-5-mini and GPT-4o top the leaderboard, and the smallest models trail (\Cref{fig:per_game_heatmap}).
However, overall strength does not decide every game.
Qwen3-1.7B, last overall, ranks first on Sheriff (bluffing under inspection), Insider (word guessing with a hidden informed player), and the Minority Game (choosing the less popular action), showing that each game tests a different ability, and no model excels at all of them.

\begin{figure}[t]
\centering
\includegraphics[width=\linewidth]{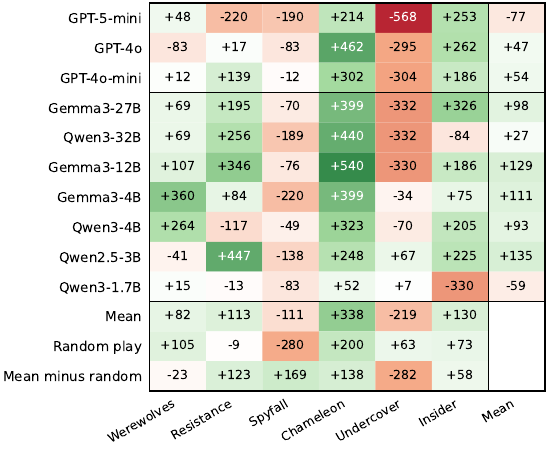}
\caption{Role gap: Elo on the minority side minus Elo on the majority side, per model and hidden-role game (green: minority stronger).
Random play: the gap when every player moves at random, which measures the game's own asymmetry.}
\label{fig:secrets_sides}
\end{figure}

\paragraph{Roles.}\label{sec:results:sides}
In hidden-role games, the two sides of a game ask for different skills.
The majority must identify opponents and coordinate, while the minority must pursue its goal without revealing its role.
A model's two ratings can differ widely (\Cref{fig:secrets_sides}), so a game-level rating hides which skill a model has.
In Resistance (forming mission teams with hidden saboteurs), GPT-5-mini plays the Resistance side best of all models (1296 Elo) but is mid-table as a Spy (1076), because it keeps Spies off the teams it proposes (\Cref{sec:results:secrets}).
Qwen2.5-3B shows the opposite gap for a different reason.
Its Spy rating is mid-table (1050), and its Resistance rating is the lowest (603).
A model can be competitive as a Spy yet struggle to identify Spies when playing for the Resistance.
Our evaluation reveals an asymmetry between performance in roles that require deception and roles that require detecting it, and its direction differs between models even within the same game.

\begin{figure}[t]
\centering
\includegraphics[width=0.85\linewidth]{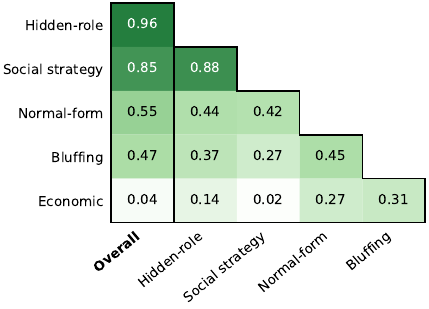}
\caption{Spearman correlation across the ten models between the Elo rankings of the game families, each fit on that family alone, and with the overall competitive ranking (first column; green: positive).
Per-family details are in \Cref{tab:subset}.}
\label{fig:family_corr}
\end{figure}

\paragraph{Game families.}
Game families rank the models differently (\Cref{fig:family_corr}, with skill and axis profiles in \Cref{sec:appendix:capability}).
Hidden-role and social strategy games agree with each other (0.88) and with the overall ranking, partly because hidden-role games supply 43\% of the competitive episodes.
An equal-weight average of the five families still reproduces the overall ranking at 0.96.
Normal-form, bluffing, and economic games reorder the models, and their differences from the overall ranking exceed what their split-half reliability allows (\Cref{tab:subset}).

The three game families that reorder the models expose weaknesses in models' strategic decision-making that the overall rating hides.
Larger models do not consistently play these games better.
In bluffing games, Gemma3-4B ranks third and Gemma3-27B last.
In Bargaining, GPT-5-mini accepts 81\% of offers that give it less than half the tokens, allowing its opponent a larger gain (\Cref{sec:results:secrets}).
Normal-form and economic Elo also track general and theory-of-mind benchmarks far less than hidden-role Elo does (\Cref{sec:results:reliability}).
We hypothesize that current training improves general reasoning and theory of mind, which help in hidden-role games, more than strategic play in these three families.

\subsection{Comparison with other evaluations}
\label{sec:results:reliability}

\begin{figure}[t]
\centering
\includegraphics[width=\linewidth]{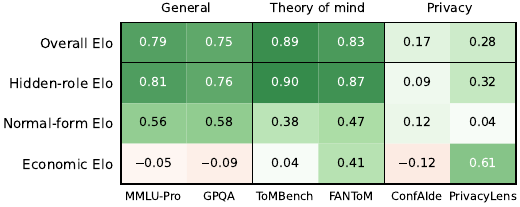}
\caption{Spearman correlation across the ten models between \socialgym ratings and outside benchmarks (green: positive; scores in \Cref{fig:benchmark_scores}).}
\label{tab:corr}
\end{figure}
To relate the social intelligence that \socialgym measures to what other benchmarks capture, we compare the family ratings with six outside benchmarks.
MMLU-Pro~\citep{wang2024mmlupro} and GPQA~\citep{rein2023gpqa} test general knowledge and reasoning.
ToMBench~\citep{chen2024tombench} and FANToM~\citep{kim2023fantom} test theory of mind with multiple-choice questions about what characters in a story or conversation know and believe.
ConfAIde~\citep{mireshghallah2024confaide} and PrivacyLens~\citep{shao2024privacylens} test whether a model recognizes when sharing private information is inappropriate.

Listed in \Cref{tab:corr}, results show that overall and hidden-role Elo performance is associated with general reasoning and theory of mind benchmark performance (Spearman $\rho$= 0.75--0.90) but is only weakly associated with performance on privacy benchmarks (0.09--0.32).
Interestingly, normal-form and economic game Elo show weaker associations with general reasoning and theory-of-mind benchmarks than hidden-role Elo does.
This suggests that \socialgym not only captures skills measured in some existing benchmarks through hidden-role games, but also captures complementary social skills not necessarily captured in benchmarks through normal-form and economic games.

Diving deeper into secret-keeping and privacy-related skills, we find that keeping information private follows a different pattern from winning.
We measure secret keeping with the rules in two games: the share of Chameleon games in which Citizens keep the secret word out of their clues, and the share of Spyfall games in which Non-Spies keep the shared location out of public discussion (\Cref{sec:results:secrets,sec:appendix:beh_private}).
Across the ten models, both measures correlate more strongly with FANToM (0.85 in Chameleon and 0.91 in Spyfall) and PrivacyLens (0.76 and 0.65) than with MMLU-Pro and GPQA (0.34 to 0.51, \Cref{fig:corr_secret}).
We see the same pattern when an LLM judge rates secret keeping in the transcripts of all six hidden-role games using the SOTOPIA-EVAL rubric~\citep{zhou2024sotopiainteractiveevaluationsocial}.
These judge scores correlate with FANToM and PrivacyLens at 0.67 and 0.77, compared with 0.10 and 0.09 for MMLU-Pro and GPQA.
By measuring both game outcomes and secret keeping during play, \socialgym assesses how well models compete and how they protect private information.

Open-ended social evaluations often use LLM judges, adding inference cost and potential evaluator bias~\citep{zheng2023judging}.
Within the same episode, SOTOPIA-EVAL scores~\citep{zhou2024sotopiainteractiveevaluationsocial} agree with game outcomes in 92.5\% of player-pair comparisons, counting judge ties as half an agreement.
At the model level, mean judge scores also closely track Elo rankings (Spearman 0.98; \Cref{sec:appendix:sotopia_eval}).
Thus, \socialgym uses verifiable game outcomes to score performance without relying on an LLM judge's interpretation of the transcript.

\subsection{Behaviors during play}
\label{sec:results:secrets}
\begin{figure}[t]
\centering
\includegraphics[width=\linewidth]{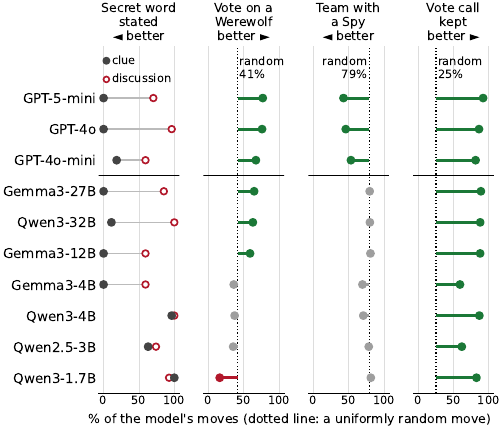}
\caption{Behavioral measures per model against a uniformly random move (dotted line).
Secret word stated: Chameleon games in which the model's Citizens give the word in the clue stage (filled) or in the discussion stage (open).
Team with a Spy: teams it proposes as a Resistance leader.
Vote call kept: its calls to vote out a player that its next vote follows.
Gray: within 10 pp of random, a threshold used only for display.}
\label{fig:behaviors}
\end{figure}

Beyond whether models win, \socialgym also allows us to examine how they play.
We specifically examine three questions:
(1) do they keep private information private, (2) can they identify hidden opponents, and (3) do their choices serve the game's goal (\Cref{fig:behaviors})?
\Cref{sec:appendix:behaviors} defines these behavioral measures and reports the results for each model.

\paragraph{Keeping information private.}
Chameleon requires Citizens to show their allies that they know the secret word without giving it to the Chameleon.
As expected, small models leak the secret word earlier and more often than large ones.
Five of the ten models state the word as their clue, Qwen3-1.7B in every game, as when all four of its Citizens give the word \emph{sleet} for the category \emph{weather event} (\Cref{fig:behaviors}).
Spyfall shows the same pattern, with six models naming the shared location outright in 19 to 82\% of their Non-Spy games.

However, even the strongest models often leak the secret in the discussion.
GPT-5-mini and GPT-4o never state the word as their clue, yet they name it in the discussion of 70\% and 96\% of their games.
In one game, GPT-5-mini Citizens hint at \emph{rugby} with \emph{scrum}, \emph{tackle}, \emph{ruck}, and \emph{lineout}, and then one says that ``the clues scrum, tackle, ruck and lineout all point to rugby.''
A caught Chameleon then guesses the word in 63\% of games, against 15\% when nobody names it.
The discussion asks players to justify their suspicions, and the models do so by showing how the clues fit the word, which reveals the word.
Even strong models fail to maintain secrecy when they shift from giving clues to explaining their suspicions, consistent with inappropriate disclosures reported in other multi-party settings~\citep{ys2026sotopiaTom,mireshghallah2024confaide}.

\paragraph{Identifying hidden opponents.}
Prior question-answering evaluations find that LLMs can struggle to infer what others know or believe~\citep{sap2022neural,kim2023fantom}.
In hidden-role games, players must make a related inference to work out who their opponents are, yet most models fail to do so reliably in multiple games.
Nine of the ten models show no reliable advantage over random selection in at least one of the six hidden-role games (\Cref{tab:appx_votes}).
These difficulties are more pronounced in the smallest models.
For example, as Villagers in Werewolves, GPT-5-mini and GPT-4o cast 77\% and 76\% of their day votes for a Werewolf, respectively, and their accuracy rises from the first day to the second as inspections and earlier votes accumulate.
By contrast, the four smallest models are not reliably better than chance, and Qwen3-1.7B casts only 16\% of its votes for a Werewolf, compared with 43\% under random voting.
In Resistance, the three GPT models propose a team with a Spy 43 to 53\% of the time, compared with 79\% under random selection, while most open models stay near chance (\Cref{fig:behaviors}).

\paragraph{Serving the game's goal.}
GPT-5-mini and GPT-4o sometimes accept payments that increase their own score but benefit their opponents more.
In Sheriff, merchants who offer a bribe carry contraband in 85\% of bags, against 29\% for those who do not.
As Sheriffs, GPT-5-mini and GPT-4o inspect bribed bags less often than the others (20 against 36\% and 29 against 69\%) and take the bribe instead.
A paid bribe gives the Sheriff more gold than an inspection does (3.96 against 3.16 on average), but a passed smuggler gains about 5 gold on the Sheriff, and the game ranks players by gold.
In Bargaining, GPT-5-mini accepts 81\% of below-half offers, whereas GPT-4o rejects 78\%, forgoing its own tokens while denying the proposer a larger gain (\Cref{sec:appendix:beh_goal}).
Qwen3-1.7B, which inspects 84\% of bribed bags, ranks first in Sheriff.
These choices suggest that models differ in how they trade off absolute payoff and relative standing, a contrast also observed in point-allocation games~\citep{ramamoorthy2025neuralfomo}.

\section{Can Inference-Time Methods Improve Play?}
\label{sec:experiments}
\label{sec:method}

Models play some roles much worse than others (\Cref{sec:results:sides}).
We ask whether an inference-time method, which changes only the prompt, can close such gaps. This would provide insight into whether LLMs carry the foundational ability to reason and act with social intelligence.
As a specific probe, we introduce \textsc{SPaRTan}, a simple method in which a model writes a strategy playbook from its own self-play, and test it against other prompt-level methods on the same games and deals.

\paragraph{Methods.}
\textsc{SPaRTan} (\textbf{S}elf-\textbf{P}lay \textbf{a}nd \textbf{R}eflect-\textbf{T}r\textbf{an}sfer) has three stages, inspired by methods that turn an agent's own trajectories into reusable insights~\citep{zhao2024expel} or into evolving context playbooks~\citep{zhang2025ace, xie2026memo}.
A model $M$ plays $N{=}10$ self-play games of each source game, reads the trajectories with their outcomes, and writes a first-person strategy playbook (prompt in \Cref{sec:appendix:reflection_prompt}).
$M$ then reads the playbook at the start of its prompt whenever it plays a target game $t$.
To test whether the advice transfers across games, we vary the games used to write the playbook.
The $1{\to}1$ and $(n{-}1){\to}1$ variants write a separate playbook for each target game $t$, from $t$ alone or from every game except $t$.
The $n{\to}n$ variant writes one playbook from all $n$ games and uses it in every target game.

We also evaluate three baselines by adding their advice or lessons to the agents' prompts.
Generic game advice is written by $M$ without seeing any game or its rules, providing a control for whether strategy advice alone improves play.
Alongside this control, we adapt two existing methods to our games.
For the Self-Refine playbook~\citep{madaan2023self}, $M$ writes a playbook for each game from its rules alone, then critiques and revises it twice without playing.
Reflexion~\citep{shinn2023reflexion} writes a lesson after each lost game and carries the last three into the next one.
We describe these adaptations and additional prompting methods in \Cref{sec:appendix:controls}, which also reports the complete results.

\paragraph{Evaluation settings.}
We evaluate these methods with GPT-5-mini and Qwen3-32B on three hidden-role games (Werewolves, Spyfall, and Resistance) due to computational constraints.
In these games, GPT-5-mini's minority-side self-play win rates are 33 to 37\% (\Cref{sec:appendix:balance}).
For each condition, we add its text to every agent on one side (the armed side), while the opposing agents use the same model without added text.
We evaluate each side separately on 60 initial deals per game and compare each condition with vanilla self-play on matching deals.
We report the change in Elo against vanilla self-play, $\Delta\mathrm{Elo} = 400\,[\log_{10}\frac{p}{1-p} - \log_{10}\frac{p_0}{1-p_0}]$, where $p$ and $p_0$ are the win rates of that side with and without the text.
We average $\Delta\mathrm{Elo}$ across the three games for each side and across both sides for the overall result, with 95\% CIs obtained by paired bootstrap over deals.
As in \Cref{subsec:elo}, \emph{alt} is the minority side and \emph{main} the majority side.

\begin{figure}[t]
\centering
\includegraphics[width=\linewidth]{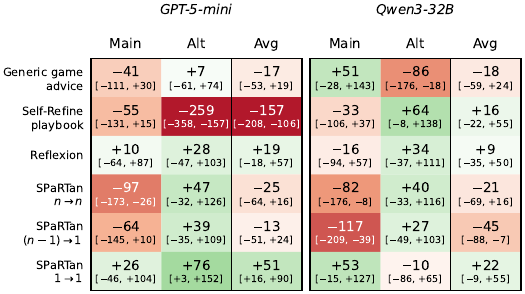}
\caption{Change in Elo against vanilla self-play, with 95\% paired-bootstrap CIs.
Avg: mean of the main and alt sides over Werewolves, Spyfall, and Resistance.}
\label{fig:spartan_delta}
\end{figure}

\begin{figure}[t]
\centering
\includegraphics[width=\linewidth]{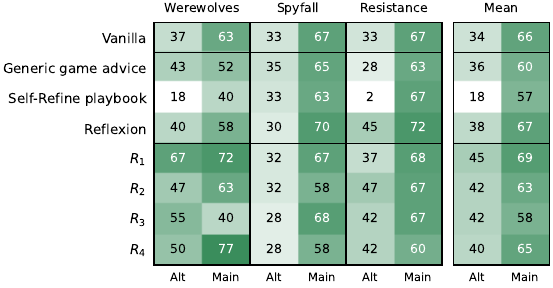}
\caption{Win rate (\%) of the armed side against vanilla GPT-5-mini, $n{=}60$ per cell.
Rows between Vanilla and $R_1$ are other prompting methods (\Cref{sec:appendix:controls}).
$R_1$ is the $1{\to}1$ playbook.
$R_2$ to $R_4$ each reflect on ten games played with the previous round's playbook.
Means pool the three games (95\% CI $\pm7$ pp).}
\label{fig:exp2:within}
\end{figure}

\paragraph{Results.}
Our findings show that the baselines yield no clear average improvement for either model.
Apart from GPT-5-mini's Self-Refine playbook, their average changes range from $-18$ to $+19$ Elo, and every 95\% CI spans zero (\Cref{fig:spartan_delta}).
GPT-5-mini's Self-Refine playbook instead lowers its average Elo by 157.
In Werewolves, it lowers the win rate of both sides, from 63 to 40\% for the Villagers and from 37 to 18\% for the Werewolves (\Cref{fig:exp2:within}).
The playbook recommends probe votes such as ``I'll vote X now to see who defends them.''
Villagers using it put 48\% of their votes on a Werewolf, against 68\% in vanilla play.
Werewolves using it both call probe votes on the first day in 95\% of games, and the Villagers put 82\% of their votes on them.
In Resistance, it tells a Spy to fail every mission that holds both Spies, which exposes both, and the Spies using the playbook win only 2\% of games.
The loss thus depends on what the playbook says, since Qwen3-32B's Self-Refine playbooks contain neither instruction and change its average by only $+16$ Elo.

In contrast, a \textsc{SPaRTan} playbook written from the target game raises GPT-5-mini's Elo by 51 on average (95\% CI $[16,90]$), with gains of 76 on the minority side and 26 on the majority side.
This gain comes mainly from Werewolves, where the playbook (\Cref{sec:appendix:reflections}) raises the Werewolves' win rate from 37 to 67\%, compared with 18 to 43\% under the three other prompting methods in \Cref{fig:exp2:within}.
However, the same approach yields little change in Spyfall or Resistance.
For Qwen3-32B, the average gain is smaller and uncertain ($+22$ Elo, 95\% CI $[-9,55]$).

Pooling experience across games provides no clear average improvement for either model.
The pooled ($n{\to}n$) and held-out ($(n{-}1){\to}1$) playbooks change GPT-5-mini's average by $-25$ and $-13$ Elo and Qwen3-32B's by $-21$ and $-45$.
Most of this loss falls on the majority side, which the pooled playbook lowers by 97 Elo for GPT-5-mini and 82 for Qwen3-32B.
Later reflection rounds ($R_2$ to $R_4$) change GPT-5-mini's average by 0 to $+19$ Elo, below the first playbook's $+51$ (\Cref{fig:spartan_delta_full}).

Self-play reflection can therefore improve play in specific settings, but the gains depend on the model, game, and role and do not consistently extend across games or reflection rounds.
Whether reinforcement learning on verifiable game outcomes can produce broader improvements remains a question for future work.

\section{Conclusion}
\label{sec:conclusion}
We introduced \socialgym, an environment of 22 multi-agent social games with verifiable outcomes, and benchmarked ten LLMs under a common tournament protocol.
Model strengths varied across social settings and roles, while overall game performance and secret keeping showed different associations with existing benchmarks.
Behavioral analyses revealed failures in protecting private information, identifying opponents, and making strategic decisions.
Self-play reflection produced gains that depended on the model and game, with no consistent benefit from pooling games or additional reflection.
\socialgym supports evaluation through verifiable outcomes without requiring LLM judges for primary scoring, and provides feedback for future training.

\section*{Limitations}

\paragraph{External validity.}
All games in \socialgym have fixed rules, fixed role structures, and rule-decided outcomes.
These properties let us score every episode without an LLM judge.
Many real social interactions, such as resolving a disagreement or building long-term trust, have no clear win/loss criterion and cannot be reduced to a single score.
Correlations with other benchmarks (\Cref{tab:corr}) describe associations across ten models.
They do not establish transfer to interactions with humans.
Whether the capabilities measured in these games carry over to such settings remains open for future work to investigate.

\paragraph{Ranking stability differs by family.}
Across random halves of the data, the economic and social-strategy rankings are the least reliable (split-half reliability 0.68 and 0.55, against 0.89 to 0.94 for the other families).
Each of these two families has only two competitive games, so conclusions about them rest on fewer reliable distinctions.

\paragraph{Diagnostic coverage.}
Per-side ratings cover hidden-role games, while behavioral measures and communication controls (\Cref{sec:results:talk}) cover selected games.
These game-specific measures do not establish a comprehensive measure of each social capability across the full suite.
Behavioral findings depend on the rules and prompts presented to the models.
In Chameleon, the rules list the Chameleon among the players in the discussion, and its own messages appear in the transcript, but the prompt does not warn Citizens that it reads theirs.

\paragraph{Elo is relative.}
The random reference player provides a fixed comparison under the tested seating and opponent conditions.
Its performance depends on how it samples legal moves and on the other players at the table, so the anchored ratings are not an absolute measure of social intelligence.
Comparisons with human players and stronger rule-based strategies remain open.

\paragraph{Scope of the inference-time experiments.}
Our inference-time experiments cover two models and three hidden-role games under self-play.
The observed gains therefore do not establish benefits across the full suite or against other opponents, and we do not evaluate methods that update model weights.

\section*{Ethics / Broader Impacts}
\socialgym measures how LLMs play social games that involve hidden roles, bluffing, persuasion, and coalition building.
Skills that win these games, such as concealing information or persuading other players, could be misused if transferred to interactions with people.
All experiments run on synthetic multi-agent games without human participants.
We release the environment and code to support research on these behaviors, and we intend them for evaluating models in games.

\section*{Acknowledgments}
We thank Michael Li, Andy Liu, Joseph Liu, and Neemesh Yadav for their valuable feedback on drafts of this paper.

This work was in part funded by the National Institute of Standards and Technology (ROR: 05xpvk416) under Federal Award ID Number 60NANB24D231 and Carnegie Mellon University (ROR: 05x2bcf33) AI Measurement Science and Engineering Center (AIMSEC).
Research supported by the NVIDIA Academic Grant Program using NVIDIA H100 GPUs on NVIDIA Brev.

\bibliography{custom}

\appendix
\crefalias{section}{appendix}
\crefalias{subsection}{appendix}

\section{Code and Data Release}
\label{sec:appendix:code}

All code and data are released at \url{https://github.com/Keyu-He/Social_Gym_Spartan}.
The repository contains: (i) the game engine layer and the 22 game implementations from \Cref{sec:benchmark}, (ii) the Elo tournament infrastructure with the Bradley--Terry fit described in \Cref{sec:appendix:elo_details}, and (iii) the \textsc{SPaRTan} pipeline scripts for roster generation, reflection generation, and per-condition evaluation.

\paragraph{Dependencies and licensing.}
The bundled \texttt{sotopia/} directory is a snapshot of an open-source dependency released under the MIT license; our use is consistent with its terms, and the contribution claimed here is limited to the game-engine layer, the games, and the experiments (see \texttt{NOTICE.md} in the repository).
Our own code, game implementations, and configuration files are released under the MIT license.
The proprietary models we evaluate (GPT-5-mini, GPT-4o, GPT-4o-mini) are accessed through their providers' APIs under the respective terms of service, and the open-weights models (Qwen3, Qwen2.5, Gemma3) are used under their published licenses.

\section{Models and Inference}
\label{sec:appendix:models}

We chose the ten models to span three providers, closed and open weights, and sizes from 1.7B to 32B.
Each open family covers a small, a mid, and a large size (Gemma3 4B, 12B, 27B; Qwen3 1.7B, 4B, 32B).
Qwen2.5-3B, the one previous-generation model, sits next to Qwen3-4B at a similar size.
\Cref{tab:models} lists the checkpoints.

We serve the seven open-weight models with vLLM 0.28 in bf16, with a 32{,}768-token context window and at most 2{,}048 new tokens per response.
We query the OpenAI models through their API, with the default reasoning effort for GPT-5-mini.
All models sample with their default generation settings.
We run Qwen3 with thinking off.
Qwen3-4B is the non-thinking Instruct-2507 release.

Each turn asks for a JSON reply that follows the game's action schema.
A reply that fails to parse counts as the model's move and is played as no action.
Requests that fail at the API level are retried, and an episode with a failed request is replayed.

\begin{table}[h]
\centering
\small
\begin{tabular}{ll}
\toprule
\textbf{Model} & \textbf{Checkpoint} \\
\midrule
GPT-5-mini & \href{https://platform.openai.com/docs/models/gpt-5-mini}{\texttt{gpt-5-mini}} (OpenAI API) \\
GPT-4o & \href{https://platform.openai.com/docs/models/gpt-4o}{\texttt{gpt-4o}} (OpenAI API) \\
GPT-4o-mini & \href{https://platform.openai.com/docs/models/gpt-4o-mini}{\texttt{gpt-4o-mini}} (OpenAI API) \\
Gemma3-27B & \href{https://huggingface.co/google/gemma-3-27b-it}{\texttt{google/gemma-3-27b-it}} \\
Gemma3-12B & \href{https://huggingface.co/google/gemma-3-12b-it}{\texttt{google/gemma-3-12b-it}} \\
Gemma3-4B & \href{https://huggingface.co/google/gemma-3-4b-it}{\texttt{google/gemma-3-4b-it}} \\
Qwen3-32B & \href{https://huggingface.co/Qwen/Qwen3-32B}{\texttt{Qwen/Qwen3-32B}} \\
Qwen3-4B & \href{https://huggingface.co/Qwen/Qwen3-4B-Instruct-2507}{\texttt{Qwen/Qwen3-4B-Instruct-2507}} \\
Qwen2.5-3B & \href{https://huggingface.co/Qwen/Qwen2.5-3B-Instruct}{\texttt{Qwen/Qwen2.5-3B-Instruct}} \\
Qwen3-1.7B & \href{https://huggingface.co/Qwen/Qwen3-1.7B}{\texttt{Qwen/Qwen3-1.7B}} \\
\bottomrule
\end{tabular}
\caption{Checkpoints of the ten models.
Open-weight checkpoints are Hugging Face ids.}
\label{tab:models}
\end{table}

\section{Elo Aggregation Details}
\label{sec:appendix:elo_details}

\paragraph{Bradley--Terry fit.}
The aggregated per-pair win/tie counts (\Cref{subsec:elo}) are fit via L2-regularized logistic regression ($C{=}0.1$, scale 400, anchored mean 1000), and 95\% confidence intervals are obtained by multinomial bootstrap on the battle counts ($n_\text{bootstrap}{=}1000$).

\paragraph{Per-game vs.\ overall ratings.}
The overall Elo (right panel of \Cref{fig:per_game_heatmap}) and per-game Elos (left panel of \Cref{fig:per_game_heatmap}) are computed by independent BT fits, each anchored at a mean of 1000. 
Overall aggregates outcomes across the union of competitive games (1{,}012 episodes per model, thousands of pairwise outcomes), while each per-game fit only sees 54--90 episodes per model.
Per-game ratings therefore have larger uncertainty and stay closer to the 1000 anchor, while Overall reflects the better-determined skill estimate.

\paragraph{Why BT-MLE rather than the online K-factor update.}
We adopt BT-MLE rather than the classical online Elo update~\citep{elo1978rating} for three reasons.
First, BT is path-independent: the rating does not depend on the order in which episodes are processed.
Second, with six episodes per pair in each game, the online update would not converge.
Third, BT is the standard for current LLM leaderboards~\citep{chiang2024chatbot}.

\paragraph{Same-score pairs.}
When two agents from different models end an episode with the same score, the pair enters the Bradley--Terry fit as a tie ($s=0.5$), that is, half a win for each side.

\paragraph{Cooperative games.}
For the four cooperation and coordination games, we report pairwise win rates separately from the competitive Elo leaderboard.
A pairwise win is a cross-model within-episode score comparison won; equal scores remain in the denominator.
This describes relative outcomes in these games, rather than the frequency of a jointly successful episode.

\begin{table}[h]
\centering
\small
\resizebox{\linewidth}{!}{
    \begin{tabular}{l|rrrr}
    \toprule
    \textbf{Model} & \textbf{BoS} & \textbf{Stag Hunt} & \textbf{Public Goods} & \textbf{Centipede} \\
    \midrule
    GPT-5-mini & 77.8 & 67.9 & 69.1 & 83.3 \\
    GPT-4o & 42.6 & 40.1 & 45.1 & 5.6 \\
    GPT-4o-mini & 40.7 & 3.1 & 42.6 & 37.0 \\
    Gemma3-27B & 40.7 & 22.8 & 61.7 & 38.9 \\
    Qwen3-32B & 40.7 & 15.4 & 19.1 & 66.7 \\
    Gemma3-12B & 44.4 & 13.6 & 46.9 & 9.3 \\
    Gemma3-4B & 51.9 & 0.0 & 0.0 & 81.5 \\
    Qwen3-4B & 38.9 & 0.0 & 10.5 & 7.4 \\
    Qwen2.5-3B & 37.0 & 56.2 & 20.4 & 42.6 \\
    Qwen3-1.7B & 22.2 & 13.0 & 21.0 & 66.7 \\
    \bottomrule
    \end{tabular}
}
\caption{Pairwise win rates (\%) on the four cooperation and coordination games, from 54 episodes per model and game.
BoS = Battle of the Sexes.
Each rate is the share of cross-model within-episode score comparisons won; equal scores remain in the denominator.}
\label{tab:coop_winrate}
\end{table}

\section{Agreement with SOTOPIA-EVAL}
\label{sec:appendix:sotopia_eval}

Open-ended social evaluations rely on LLM judges, so we compare judge scores with game outcomes and model rankings.
We scored every seat of all 4{,}588 tournament episodes on the seven SOTOPIA-EVAL dimensions~\citep{zhou2024sotopiainteractiveevaluationsocial} with Qwen3.8-27B, which reads the full transcript.
The overall score is the mean of the seven dimension scores.

For the 18 competitive games, we compare every pair of players with different rewards in the same episode.
A pair receives 1 if the player with the higher reward has the higher judge score, 0.5 if the judge scores are tied, and 0 otherwise.
The average across 15{,}501 pairs is 92.5\%.
Using the mean of the six dimensions other than goal achievement gives 81.0\% agreement.
We pool all pairs, so episodes with more pairs receive more weight.
The judge sees the game outcomes in the transcript.

\begin{figure*}[t]
\centering
\includegraphics[width=\textwidth]{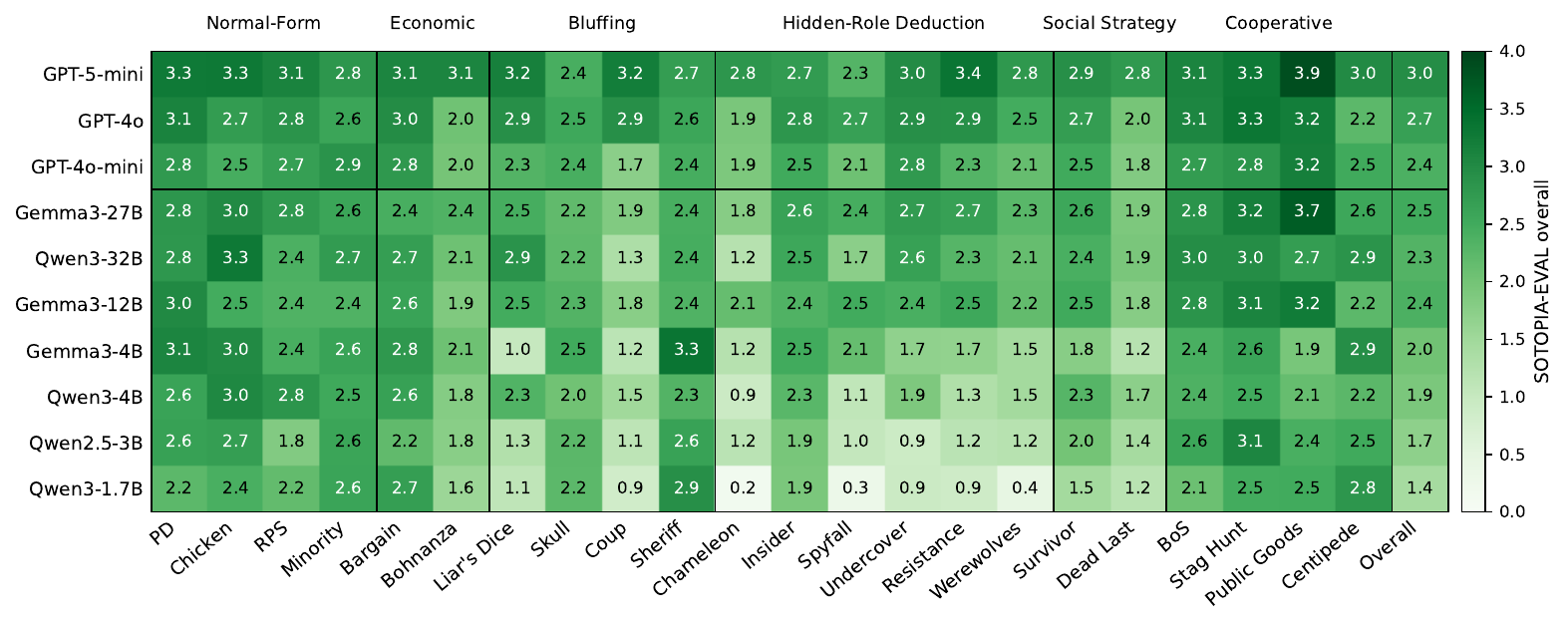}
\caption{SOTOPIA-EVAL overall score (mean of its seven dimensions, judged by Qwen3.8-27B from the full transcript) of each model in each game, every seat scored.
Game order and dividers follow \Cref{fig:per_game_heatmap}, with the four cooperative games added; the rightmost column averages all 22 games.}
\label{fig:sotopia_heatmap}
\end{figure*}

\begin{figure}[t]
\centering
\includegraphics[width=\linewidth]{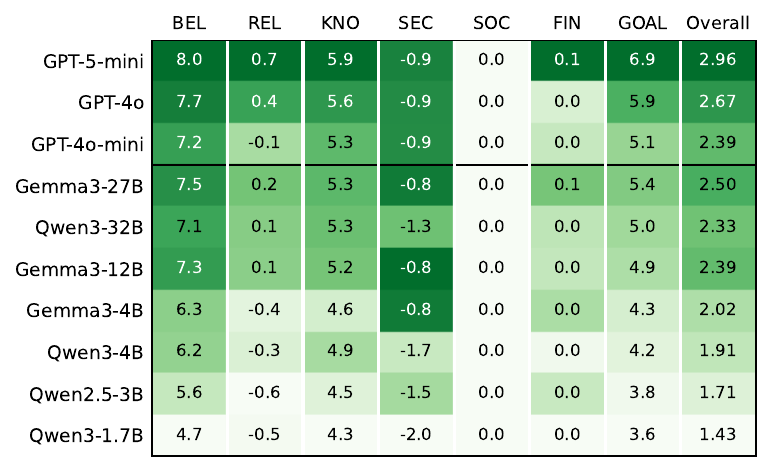}
\caption{The seven SOTOPIA-EVAL dimensions (believability, knowledge and goal from 0 to 10; relationship and financial benefits from $-5$ to 5; secret and social rules from $-10$ to 0, with 0 best) per model over all 22 games, judged by Qwen3.8-27B; cells are shaded within each column.
Social rules is 0 almost everywhere, since in-game deception does not count as a violation.}
\label{fig:sotopia_dims}
\end{figure}

Across the ten models, mean overall scores have a Spearman correlation of 0.98 with Elo on the 18 competitive games.
When we exclude goal achievement, the model-level correlation is 0.95.
\Cref{fig:sotopia_heatmap,fig:sotopia_dims} show the Qwen3.8 scores.

\section{Capability-Profile Elo}
\label{sec:appendix:capability}

This appendix disaggregates the overall leaderboard into capability profiles along the taxonomy axes of \Cref{tab:taxonomy}.
For each group of games (a category, a skill tag, or a structural axis) we refit the Bradley--Terry model on the pooled pairwise outcomes of that group's games only, with hyperparameters and bootstrap identical to \Cref{sec:appendix:elo_details}.
We refit per group rather than averaging per-game Elos because the per-game fits carry heteroscedastic uncertainty.
Only the 18 competitive games enter these fits (cooperative games are scored by win rate, \Cref{tab:coop_winrate}).
Skill tags overlap, so these axes are not independent measurements.

\begin{figure*}[t]
\centering
\includegraphics[width=\textwidth]{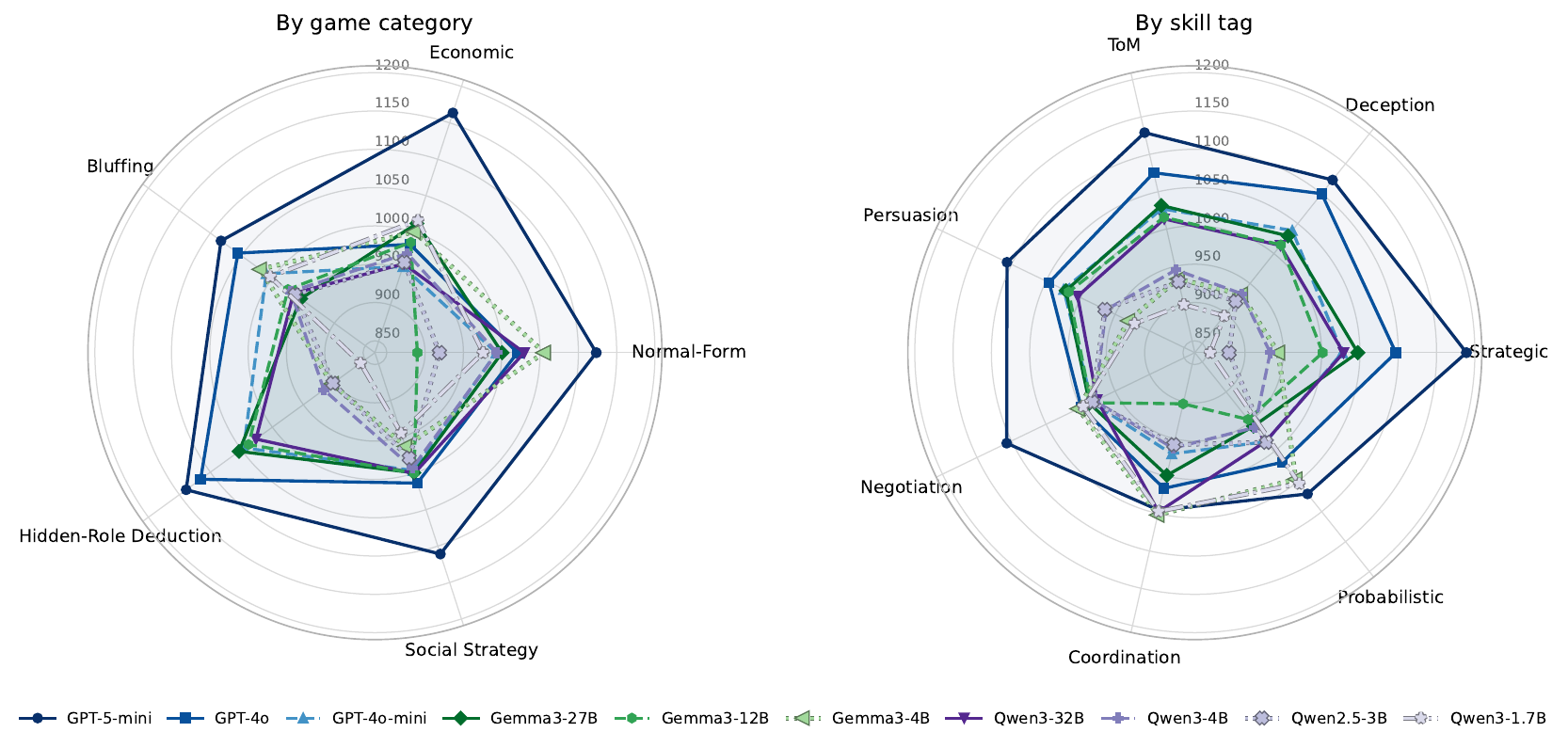}
\caption{Capability-profile Elo: independent Bradley--Terry refits per game category (left) and per skill tag (right; tags from \Cref{tab:taxonomy}), with the same hyperparameters and bootstrap as the overall fit.
Both panels share the same radial scale.}
\label{fig:capability_radar}
\end{figure*}

\paragraph{Findings.}
Three patterns stand out in \Cref{fig:capability_radar} (full numbers in \Cref{tab:capability_categories,tab:capability_skills}).
First, the six hidden-role deduction games alone reproduce the overall ranking.
Of the 45 pairings among our ten models, the full 18-game tournament separates 38 at 95\% confidence (non-overlapping bootstrap CIs).
A fit on the six hidden-role games orders 42 of the 45 pairs as the full fit does (Spearman 0.96) and separates 35 pairs, all in the full-fit order, from 43\% of the tournament's episodes (1{,}620 of 3{,}778).
Second, bluffing preserves only the top of the ranking.
GPT-5-mini and GPT-4o remain ranks 1 and 2, but Gemma3-4B rises to third and Gemma3-27B falls to last; 6 of the 27 pairs this fit separates run against the full-fit order.
Third, normal-form and economic games reorder everything below the leader.
On normal-form games, Gemma3-4B is second and Gemma3-12B last, 166 Elo apart.
The fit separates 26 pairs, 5 of them in the reverse of the full-fit order, so matrix games separate models in an order of their own.
The two economic games (Bargaining and Bohnanza) separate GPT-5-mini from every other model and no other pair; the remaining nine sit within 64 Elo, with Qwen3-1.7B, the smallest model, second.
Their ranking is uncorrelated with the full one (Spearman 0.04).
On the skill axes, GPT-5-mini leads every tag except Coordination, and Deception orders 43 of the 45 pairs as the full fit does.
Negotiation is the most compressed tag: the nine models below GPT-5-mini lie within 27 Elo, with no pair among them separated.
Coordination reorders the field, with Gemma3-4B, Qwen3-32B, Qwen3-1.7B and GPT-5-mini within 6 Elo at the top and Gemma3-12B last (Spearman 0.10).
Reporting ratings by family and by skill tag therefore shows abilities that the overall rating, which closely follows the hidden-role games, does not capture.

\begin{table*}[t]
\centering
\small
\resizebox{\textwidth}{!}{
\begin{tabular}{l|ccccc}
    \toprule
    \textbf{Model} & \textbf{Normal-Form (4)} & \textbf{Economic (2)} & \textbf{Bluffing (4)} & \textbf{Hidden-Role (6)} & \textbf{Social Strategy (2)} \\
    \midrule
    GPT-5-mini  & \textbf{1123} {\scriptsize [1095, 1151]} & \textbf{1163} {\scriptsize [1118, 1214]} & \textbf{1082} {\scriptsize [1061, 1101]} & \textbf{1139} {\scriptsize [1122, 1154]} & \textbf{1110} {\scriptsize [1090, 1131]} \\
    GPT-4o      & 1020 {\scriptsize [990, 1046]} & 983 {\scriptsize [942, 1025]} & 1055 {\scriptsize [1035, 1076]} & 1115 {\scriptsize [1100, 1131]} & 1013 {\scriptsize [995, 1032]} \\
    GPT-4o-mini & 992 {\scriptsize [966, 1022]} & 952 {\scriptsize [911, 992]} & 1009 {\scriptsize [990, 1028]} & 1046 {\scriptsize [1031, 1060]} & 996 {\scriptsize [978, 1014]} \\
    Gemma3-27B  & 1000 {\scriptsize [971, 1027]} & 1011 {\scriptsize [971, 1052]} & 954 {\scriptsize [935, 972]} & 1053 {\scriptsize [1038, 1068]} & 999 {\scriptsize [981, 1017]} \\
    Qwen3-32B   & 1029 {\scriptsize [1001, 1058]} & 956 {\scriptsize [914, 996]} & 965 {\scriptsize [948, 984]} & 1026 {\scriptsize [1009, 1042]} & 1000 {\scriptsize [981, 1017]} \\
    Gemma3-12B  & 889 {\scriptsize [861, 917]} & 985 {\scriptsize [945, 1025]} & 974 {\scriptsize [956, 994]} & 1038 {\scriptsize [1023, 1053]} & 999 {\scriptsize [980, 1019]} \\
    Gemma3-4B   & 1055 {\scriptsize [1026, 1081]} & 1000 {\scriptsize [959, 1039]} & 1019 {\scriptsize [998, 1040]} & 903 {\scriptsize [888, 918]} & 963 {\scriptsize [944, 982]} \\
    Qwen3-4B    & 993 {\scriptsize [966, 1019]} & 970 {\scriptsize [929, 1010]} & 970 {\scriptsize [950, 988]} & 916 {\scriptsize [901, 931]} & 992 {\scriptsize [975, 1012]} \\
    Qwen2.5-3B  & 918 {\scriptsize [890, 947]} & 958 {\scriptsize [918, 1001]} & 963 {\scriptsize [945, 981]} & 902 {\scriptsize [886, 916]} & 979 {\scriptsize [960, 998]} \\
    Qwen3-1.7B  & 975 {\scriptsize [951, 1003]} & 1016 {\scriptsize [976, 1054]} & 1003 {\scriptsize [986, 1021]} & 857 {\scriptsize [842, 873]} & 944 {\scriptsize [926, 963]} \\
    \bottomrule
    \end{tabular}
}
\caption{Per-category Bradley--Terry Elo (95\% bootstrap CIs in brackets, $n_\text{bootstrap}{=}1000$); rows sorted by overall Elo, bold marks each column's leader, and parenthesized counts are the number of games per category.}
\label{tab:capability_categories}
\end{table*}

\begin{table*}[t]
\centering
\small
\resizebox{\textwidth}{!}{
    \begin{tabular}{l|ccccccc}
    \toprule
    \textbf{Model} & \textbf{Strategic (8)} & \textbf{Deception (10)} & \textbf{ToM (11)} & \textbf{Persuasion (5)} & \textbf{Negotiation (5)} & \textbf{Coordination (3)} & \textbf{Probabilistic (2)} \\
    \midrule
    GPT-5-mini  & \textbf{1188} {\scriptsize [1170, 1208]} & \textbf{1122} {\scriptsize [1109, 1136]} & \textbf{1129} {\scriptsize [1117, 1141]} & \textbf{1106} {\scriptsize [1090, 1122]} & \textbf{1107} {\scriptsize [1090, 1124]} & 1045 {\scriptsize [1019, 1071]} & \textbf{1070} {\scriptsize [1041, 1104]} \\
    GPT-4o      & 1096 {\scriptsize [1080, 1113]} & 1100 {\scriptsize [1086, 1111]} & 1075 {\scriptsize [1065, 1086]} & 1045 {\scriptsize [1030, 1060]} & 998 {\scriptsize [982, 1015]} & 1016 {\scriptsize [991, 1039]} & 1017 {\scriptsize [986, 1046]} \\
    GPT-4o-mini & 1026 {\scriptsize [1008, 1044]} & 1038 {\scriptsize [1025, 1051]} & 1027 {\scriptsize [1016, 1038]} & 1024 {\scriptsize [1010, 1038]} & 983 {\scriptsize [969, 1000]} & 969 {\scriptsize [945, 995]} & 983 {\scriptsize [951, 1013]} \\
    Gemma3-27B  & 1047 {\scriptsize [1030, 1063]} & 1029 {\scriptsize [1017, 1041]} & 1031 {\scriptsize [1021, 1041]} & 1021 {\scriptsize [1007, 1036]} & 987 {\scriptsize [971, 1002]} & 998 {\scriptsize [974, 1024]} & 957 {\scriptsize [927, 988]} \\
    Qwen3-32B   & 1028 {\scriptsize [1012, 1046]} & 1013 {\scriptsize [1002, 1025]} & 1013 {\scriptsize [1002, 1024]} & 1004 {\scriptsize [989, 1018]} & 976 {\scriptsize [959, 991]} & 1047 {\scriptsize [1022, 1074]} & 982 {\scriptsize [948, 1011]} \\
    Gemma3-12B  & 1001 {\scriptsize [984, 1018]} & 1014 {\scriptsize [1002, 1026]} & 1015 {\scriptsize [1004, 1026]} & 1017 {\scriptsize [1002, 1033]} & 985 {\scriptsize [967, 1001]} & 902 {\scriptsize [876, 928]} & 945 {\scriptsize [916, 973]} \\
    Gemma3-4B   & 943 {\scriptsize [927, 960]} & 931 {\scriptsize [919, 944]} & 933 {\scriptsize [921, 943]} & 930 {\scriptsize [915, 944]} & 1003 {\scriptsize [988, 1019]} & \textbf{1051} {\scriptsize [1026, 1077]} & 1045 {\scriptsize [1016, 1075]} \\
    Qwen3-4B    & 932 {\scriptsize [915, 948]} & 932 {\scriptsize [919, 945]} & 945 {\scriptsize [934, 955]} & 963 {\scriptsize [950, 977]} & 979 {\scriptsize [962, 994]} & 962 {\scriptsize [936, 988]} & 959 {\scriptsize [930, 987]} \\
    Qwen2.5-3B  & 879 {\scriptsize [862, 897]} & 920 {\scriptsize [909, 932]} & 929 {\scriptsize [919, 940]} & 964 {\scriptsize [949, 978]} & 982 {\scriptsize [967, 998]} & 957 {\scriptsize [930, 983]} & 984 {\scriptsize [955, 1013]} \\
    Qwen3-1.7B  & 854 {\scriptsize [837, 872]} & 896 {\scriptsize [883, 907]} & 898 {\scriptsize [888, 909]} & 921 {\scriptsize [907, 936]} & 996 {\scriptsize [980, 1012]} & 1047 {\scriptsize [1022, 1071]} & 1053 {\scriptsize [1027, 1083]} \\
    \bottomrule
    \end{tabular}
}
\caption{Per-skill-tag Bradley--Terry Elo (95\% bootstrap CIs, $n_\text{bootstrap}{=}1000$); rows sorted by overall Elo and bold marks each column's leader.
Tags come from \Cref{tab:taxonomy} (parenthesized counts: games per tag), so a game contributes to every tag it carries; Cooperation is omitted because Prisoner's Dilemma is the only competitive game that carries it.}
\label{tab:capability_skills}
\end{table*}

\begin{figure*}[t]
\centering
\includegraphics[width=\textwidth]{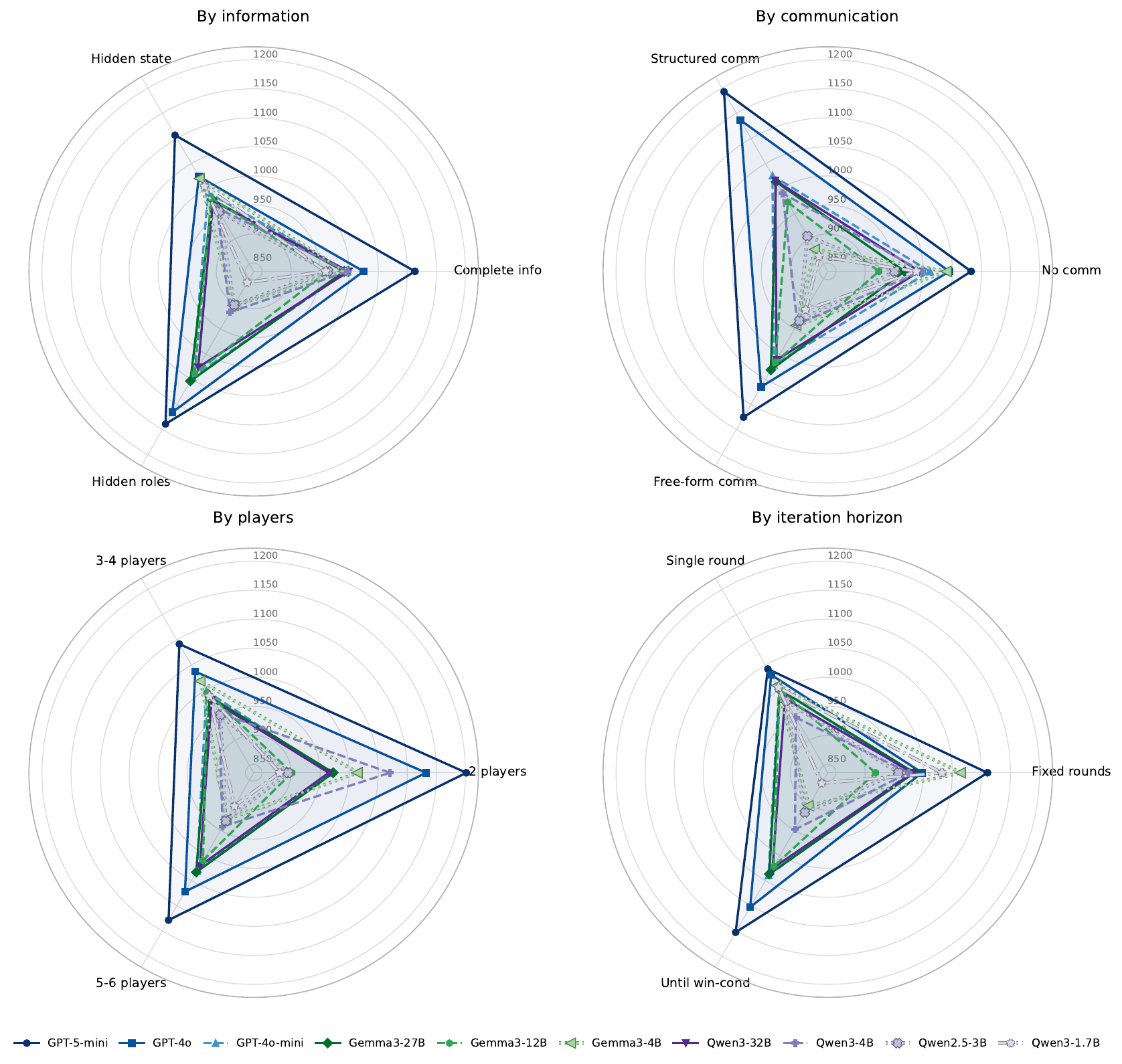}
\caption{Bradley--Terry Elo refit along four structural axes from \Cref{tab:taxonomy}: information structure, communication mode, player count, and iteration horizon.
All panels share the same radial scale.}
\label{fig:capability_axes}
\end{figure*}

\paragraph{Structural axes.}
\Cref{fig:capability_axes} repeats the refit along four structural axes of \Cref{tab:taxonomy}: information structure, communication mode, player count, and iteration horizon.
We find two patterns.
First, hidden roles split the field by strength.
Each of the six highest-rated models gains Elo from complete information to hidden roles; GPT-4o rates 1024 on both complete-information and hidden-state games, then rises to 1115 on hidden-role games.
Each of the four models with 4B parameters or fewer loses Elo: Qwen2.5-3B ends 61 lower (963 $\to$ 954 $\to$ 902) and Qwen3-1.7B 102 lower (959 $\to$ 1004 $\to$ 857).
Second, separation grows with interaction horizon: the standard deviation of the ten models' Elos rises from 26 on single-round games to 53 on fixed-round games to 89 on open-ended games.
Only the open-ended bucket separates most model pairs in the full-fit order (37 of 45, versus 16 on fixed-round and 10 on single-round games); the fixed-round fit separates 28 pairs, 12 of them in the reverse order.
Thus, open-ended games provide the clearest separation consistent with the overall ranking.
The communication and player-count axes track the game families: hidden-role games make up six of the ten free-form dialogue games and five of the eight games with five or six players, which follow the full ranking (Spearman 0.98 and 0.95), while normal-form games make up most of the games without communication and the two-player games, which reorder it (0.59 and 0.69).

\section{Game Engine Details}
\label{sec:appendix:engine}

This appendix documents the game engine behind the 22 games (\Cref{subsec:sys_arch}).
All games run on one shared interaction loop.
Each game implements a small Python script, along with a JSON configuration file and a JSON roster file that assigns models to seats and roles.
The script holds three thin components: an environment subclass that overrides engine hooks (e.g., reset and elimination checks), an \texttt{ActionHandler} that parses action arguments and updates the game's internal state, and a rule-based end evaluator.
Adding a game requires no engine changes; the engine contains no game-specific code.

\paragraph{Game configuration.}
The configuration declares the FSM and the role-level information (the assignment of agents to roles and teams comes from the per-episode roster): an initial state; a transition map from each state to its successor; and per-state properties specifying which roles act (\texttt{acting\_roles}; all alive agents if unspecified), which of the five action types are legal (\texttt{speak}; \texttt{action}, whose free-form argument the \texttt{ActionHandler} parses; \texttt{non-verbal communication}; \texttt{none}; \texttt{leave}), the turn scheduler (\texttt{simultaneous}, \texttt{round-robin}, or \texttt{random}), and the message visibility scope (\texttt{public}, \texttt{team}, or \texttt{private}).
The configuration also carries the role goals and role secrets injected into each agent's prompt; some games (e.g., Werewolves) declare their end conditions in the configuration as well, while others implement them directly in the evaluator.
\Cref{fig:werewolves_fsm} shows the Werewolves FSM as declared by its configuration.

\begin{figure}[t]
\centering
\begin{tikzpicture}[
  node distance=3.5mm,
  state/.style={rectangle, rounded corners=3pt, draw=black!70, align=center, inner sep=5pt, text width=3.7cm, font=\scriptsize},
  night/.style={state, fill=black!8},
  day/.style={state, fill=black!2},
  arr/.style={-{Stealth[length=2mm]}, thick}
]
\node[night] (nw) {\textbf{\small Night\_werewolf}\\[2pt] acts: Werewolf\\ scheduler: round-robin\\ visibility: team};
\node[night, below=of nw] (ns) {\textbf{\small Night\_seer}\\[2pt] acts: Seer\\ scheduler: round-robin\\ visibility: private};
\node[night, below=of ns] (nwi) {\textbf{\small Night\_witch}\\[2pt] acts: Witch\\ scheduler: round-robin\\ visibility: private};
\node[day, below=of nwi] (dd) {\textbf{\small Day\_discussion}\\[2pt] acts: all alive (speak)\\ scheduler: round-robin\\ visibility: public};
\node[day, below=of dd] (dv) {\textbf{\small Day\_vote}\\[2pt] acts: all alive (vote)\\ scheduler: simultaneous\\ visibility: public};
\draw[arr] (nw) -- (ns);
\draw[arr] (ns) -- (nwi);
\draw[arr] (nwi) -- (dd);
\draw[arr] (dd) -- (dv);
\draw[arr] (dv.west) .. controls +(-1.1,0.4) and +(-1.1,-0.4) .. (nw.west);
\end{tikzpicture}
\caption{The Werewolves FSM: five states in a fixed cycle, each annotated with which roles act, the turn scheduler, and the message visibility scope.
States and visibility scopes are declared in the JSON configuration; the three night states leave the scheduler unspecified and inherit the environment's round-robin default.
End conditions are checked every turn under a 40-turn cap: Villagers win when no werewolves remain; Werewolves win when they reach parity with the village.}
\label{fig:werewolves_fsm}
\end{figure}

\paragraph{Scheduler and state transitions.}
The scheduler (\Cref{alg:scheduler}) advances the FSM automatically.
In a \texttt{simultaneous} state, every eligible agent acts in the same turn and the state transitions after that turn; in \texttt{round-robin} and \texttt{random} states, one eligible agent acts per turn and the state transitions after $N$ turns, where $N$ is the number of eligible agents.
Eligibility is recomputed every turn: an agent is eligible if it is alive and, when the state restricts actors, its role is listed in \texttt{acting\_roles}; eliminated agents are masked to the no-op action and drop out of the count.
A transition resets the state's turn counter and the round-robin pointer.
The configuration map assigns each state exactly one successor (possibly itself); games whose flow branches on outcomes, such as Resistance's mission votes, implement the branch in the game subclass, either by setting the next state directly or by overriding the transition hook.

\begin{figure*}[t]
\begin{minipage}[t]{0.48\textwidth}
\begin{algorithm}[H]
\caption{Turn scheduling and state transition}
\label{alg:scheduler}
\begin{algorithmic}[1]
\State $P \gets \textsc{StateProps}(s)$ \Comment{$s$: current FSM state}
\State $E \gets$ alive agents, filtered by $P.\text{roles}$ if set
\If{$P.\text{scheduler} = \texttt{simultaneous}$}
  \State unmask every $a \in E$;\; $T \gets 1$
\ElsIf{$P.\text{scheduler} = \texttt{round-robin}$}
  \State unmask the next agent of $E$ cyclically
  \State $T \gets |E|$
\Else \Comment{\texttt{random}}
  \State unmask one random $a \in E$;\; $T \gets |E|$
\EndIf
\State \emph{(after the turn)} $t_s \gets t_s + 1$
\If{$t_s \geq T$}
  \State $s \gets \text{transition}[s]$;\; reset counters
\EndIf
\end{algorithmic}
\end{algorithm}
\end{minipage}\hfill
\begin{minipage}[t]{0.48\textwidth}
\begin{algorithm}[H]
\caption{Visibility filtering when delivering message $m$ from sender $u$, with optional receiver list $R$}
\label{alg:visibility}
\begin{algorithmic}[1]
\Procedure{Deliver}{$u$, $m$, $R$}
\State $v \gets \textsc{StateProps}(s).\text{visibility}$
\ForAll{agents $a$}
  \If{$R$ is specified}
    \State $\textit{see} \gets (a \in R)$ \Comment{targeted}
  \ElsIf{$u = \text{Environment}$}
    \State $\textit{see} \gets \textbf{true}$
  \ElsIf{$v = \texttt{public}$}
    \State $\textit{see} \gets \textbf{true}$
  \ElsIf{$v = \texttt{team}$}
    \State $\textit{see} \gets (\text{team}(u) = \text{team}(a))$
  \Else \Comment{\texttt{private}}
    \State $\textit{see} \gets (u = a)$
  \EndIf
  \If{\textit{see}} append $m$ to $\text{buffer}[a]$ \EndIf
\EndFor
\EndProcedure
\end{algorithmic}
\end{algorithm}
\end{minipage}
\end{figure*}

\paragraph{Visibility filtering.}
Every message has one of three scopes.
Public messages reach all alive agents, such as day discussion and vote results.
Team messages reach one faction, such as the werewolves' night-phase kill votes.
Private messages reach one agent, such as role cards and Seer inspections.
Messages are filtered at delivery time (\Cref{alg:visibility}).
Each agent holds a private message buffer, and a message is appended to a buffer only if that agent may see it under the current state's visibility scope.
Environment messages are public unless explicitly targeted at a receiver list; targeted delivery is how private information, such as a Seer inspection result, reaches exactly one agent.
An agent's observation each turn is the flush of its own buffer, so an agent never observes a message it was not entitled to see.

\paragraph{End evaluation and rewards.}
The end evaluator runs after every turn and checks the game's rule-based end conditions.
For example, in Werewolves these are team elimination and parity: the village wins once both werewolves are eliminated, and the werewolves win once they are at least as numerous as the surviving village team (Seer and Witch included).
On termination it emits a rule-decided outcome score for every agent, and the tournament pipeline derives pairwise wins and losses by comparing these scores within an episode (\Cref{subsec:elo}).
The prompt template through which agents observe the game is in \Cref{sec:appendix:gameplay_prompt}.

\section{Game-Play Prompt}
\label{sec:appendix:gameplay_prompt}

Each agent receives a prompt at every turn it is asked to act.
The template is shared across all 22 games; per-game variation lives in the \texttt{description}, action vocabulary, and format instructions.
When an agent's roster entry has \texttt{include\_reflection: true}, the playbook generated by the reflection step (\Cref{sec:appendix:reflection_prompt}) is loaded once and prepended via the \texttt{\{reflection\}} slot; otherwise the slot is empty.

\begin{quote}
\small\ttfamily
\noindent\{reflection\}\\[6pt]
Imagine you are playing the game as \{agent\}.\\[4pt]
Here is the description of the game: \{description\}\\[4pt]
Your (\{agent\}'s) goal: \{goal\}\\
\{secret\}\\[4pt]
Here is the context of the interaction:\\
\{history\}\\[4pt]
Your available action type(s): [\{action\_list\}].\\
\{action\_instructions\}\\[6pt]
Please only generate a JSON string including the action type and the argument.\\
Your action should follow the given format:\\
\{format\_instructions\}
\end{quote}

\paragraph{Slot semantics.}
\begin{itemize}
    \setlength\itemsep{0pt}
    \item \texttt{\{reflection\}}: full text of the playbook (e.g.\ $R_t$ for some round $t$), or empty for the vanilla baseline.
    \item \texttt{\{agent\}}: the agent's display name (e.g.\ ``Stephen'').
    \item \texttt{\{description\}}: the game-specific scenario string (e.g.\ Werewolves' phase rules and win conditions).
    \item \texttt{\{goal\}}: the role-conditioned goal (e.g.\ ``Identify werewolves'' for a Villager).
    \item \texttt{\{secret\}}: any private information the role is given at game start (Werewolf identities, Spyfall location, etc.); empty for roles without secrets.
    \item \texttt{\{history\}}: visible message log filtered by the partial-observability layer (see \Cref{subsec:sys_arch}).
    \item \texttt{\{action\_list\}}, \texttt{\{action\_instructions\}}, \texttt{\{format\_instructions\}}: the actions available in the current FSM state and the JSON schema the agent must output.
\end{itemize}

\section{Reflection Prompt}
\label{sec:appendix:reflection_prompt}

The reflection model is given the full trajectories of $N$ self-play games and asked to produce a first-person strategic playbook.
For iterated rounds ($t \geq 2$) the prompt additionally includes the previous-round playbook $R_{t-1}$ and asks the model to revise rather than rewrite.
The exact prompt template is:

\begin{quote}
\small\ttfamily
\noindent\{prior\_section\}Below are the full trajectories of all \{num\_games\} games, showing every player's actions and the outcome:\\[2pt]
\{game\_summaries\}\\[6pt]
Write an \textbf{internal monologue} of transferable social reasoning skills. Frame your insights around general capabilities that apply across many social games, such as:\\[2pt]
- \textbf{Deception}: hiding your true intentions, bluffing, maintaining a consistent false persona\\
- \textbf{Detection}: identifying when others are lying, spotting inconsistencies, reading behavioral patterns\\
- \textbf{Persuasion}: convincing others to act in your interest, building credibility, framing arguments\\
- \textbf{Information management}: when to reveal, withhold, or fabricate information; timing of disclosures\\
- \textbf{Coalition dynamics}: building alliances, breaking enemy alliances, knowing when to lead vs follow\\
- \textbf{Timing and patience}: when to act early vs wait, when to commit vs stay flexible\\[6pt]
Requirements:\\
- Write in first person (``I should\ldots'', ``When I need to hide information\ldots'', ``A pattern I noticed is\ldots'')\\
- Derive insights from the games above, but write the rules so they apply beyond any specific game\\
- Focus on actionable lessons, not abstract observations\\
- Do NOT reference specific game numbers (e.g., ``Game 3'', ``Games 5-8''). Your future self will not have access to these transcripts, so such references would be meaningless\\[6pt]
This monologue will be prepended to your system prompt in future social games.
Write it so that reading it once before any social strategy game will meaningfully improve your play.
\end{quote}

The \texttt{prior\_section} for $t = 1$ is:
\begin{quote}
\small\ttfamily
You just played \{num\_games\} self-play games (\{games\_desc\}).
Each player was an independent instance of you and only saw its own role's private information; the transcripts below reveal the hidden moves of every role, but the players themselves did not have this view during play.
\end{quote}
For $t \geq 2$ it instead reads:
\begin{quote}
\small\ttfamily
You previously wrote the following strategic playbook for yourself:\\[2pt]
-{}-{}- YOUR PREVIOUS PLAYBOOK -{}-{}-\\
\{prior\_reflection\}\\
-{}-{}- END PREVIOUS PLAYBOOK -{}-{}-\\[2pt]
You then played \{num\_games\} more games using this playbook (\{games\_desc\}). Each player was an independent instance of you and only saw its own role's private information.\\
Based on these new games, \textbf{revise your playbook}.
Keep rules that worked, remove or modify rules that didn't help, and add new insights.
Output the complete revised playbook (not just the changes).
\end{quote}

\paragraph{Design choices.}
(i) The first-person framing is meant to encourage the model to treat the playbook as advice to itself, so the playbook is phrased as instructions for later play.
(ii) Forbidding game-number references prevents the model from writing rules that would be uninterpretable at deployment, when only the playbook (not the source transcripts) is available.

\section{Prompting Methods}
\label{sec:appendix:controls}

\begin{figure}[t]
\centering
\includegraphics[width=\linewidth]{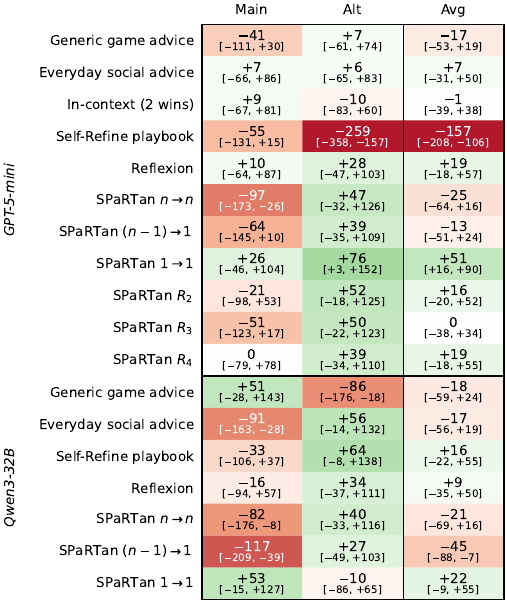}
\caption{Change in Elo against vanilla self-play for every prompting method, with 95\% paired-bootstrap CIs.
$R_2$ to $R_4$ are the later reflection rounds of the $1{\to}1$ playbook.
Qwen3-32B writes its own generic game advice and everyday social advice.}
\label{fig:spartan_delta_full}
\end{figure}

\begin{figure}[t]
\centering
\includegraphics[width=\linewidth]{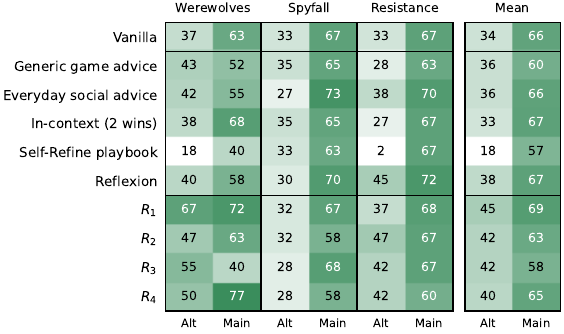}
\caption{Win rate (\%) of the armed side against vanilla GPT-5-mini for every prompting method, $n{=}60$ per cell.
Means pool the three games.}
\label{fig:within_full}
\end{figure}

Every method puts its text in the same slot as a \textsc{SPaRTan} playbook, whose $R_1$ versions are 1,062 to 1,511 words long; the methods below produce texts of similar length.
The text is prepended to the prompt of every agent on the armed side, and the other side uses the same model without added text.
\Cref{fig:spartan_delta} and \Cref{fig:exp2:within} both report generic game advice, the Self-Refine playbook, and Reflexion.
\Cref{fig:spartan_delta_full} and \Cref{fig:within_full} give the results for every method.
Each condition is paired with vanilla self-play on matching initial deals, with 60 deals per game and side before exclusions.
For Qwen3-32B's generic game advice, excluding deals with a request failure in either condition leaves 55 to 58 paired deals per game and side.

\paragraph{Generic game advice.}
GPT-5-mini receives the writing instructions of the reflection prompt (\Cref{sec:appendix:reflection_prompt}) with no trajectory and no game rules, and writes 1,437 words of advice from its prior knowledge of social games.
Qwen3-32B follows the same procedure to write its own advice.
We use the same text in every game.
The prompt asks for 1,200 words and drops the two requirements that refer to the games (deriving insights from them and not citing game numbers); the rest is identical to the reflection prompt.

\paragraph{Everyday social advice.}
GPT-5-mini receives the same request, but about everyday social life, with game tactics excluded.
The 1,617-word output is also shared across games.
Qwen3-32B also writes its own advice using this prompt, producing 1,286 words shared across games.
The prompt is:
\begin{quote}
\small\ttfamily
Write about 1400 words of an \textbf{internal monologue} about how I handle everyday social situations well. Frame it around general interpersonal skills such as: Rapport and trust, Listening, Feedback, Disagreement, Negotiating at work, Groups, Repair.\\[6pt]
\textbf{Requirements:}\\
- Write in first person (``I should\ldots'', ``When someone\ldots'', ``A habit I keep is\ldots'')\\
- Focus on actionable lessons, not abstract observations\\
- Write about ordinary social life, not games: do not mention bluffing, deception, hidden roles, voting, alliances, or winning\\
This monologue will be prepended to your system prompt in future social games.
Write it so that reading it once before any social strategy game will meaningfully improve your play.
\end{quote}

\paragraph{In-context trajectories.}
For each game we run 20 vanilla self-play games, disjoint from the evaluation deals, and give the armed side the first two games its side won.
The text states the game rules and then the two complete game records, formatted as the reflector sees them in \Cref{sec:appendix:reflection_prompt} (719 to 4,271 words).
It opens with: \texttt{Below are 2 complete records of earlier games of \{game\} that the \{side\} side won.
Study how that side played before choosing your own actions.}

\paragraph{Reflexion.}
We adapt Reflexion~\citep{shinn2023reflexion} to our self-play protocol.
The armed side plays the 60 evaluation games in sequence.
After each game it loses, GPT-5-mini reads the rules and the record of that game and writes a short lesson, and before each game the armed agents receive the last three lessons of their chain under the header \texttt{Lessons from your previous games on this side, most recent last}.
Qwen3-32B writes its own lessons in the same way.
Unlike \textsc{SPaRTan}, which reflects once on ten self-play games before evaluation and keeps the resulting playbook fixed, Reflexion updates its memory during evaluation from one lost game at a time, and each lesson is a few sentences long.
Reflexion retries the same task instance; here every game is a new deal of the same game.
Because each game depends on the lessons from earlier ones, the 60 games of a Reflexion chain are not independent.
The lesson prompt is:
\begin{quote}
\small\ttfamily
You are playing the game \{game\} on the \{side\} side. Below are the rules and the record of a game you just played, which you \{outcome\}.\\[2pt]
\{rules\}\\
\{record\}\\[6pt]
In a few sentences, diagnose the decisive mistakes or moves that led to this outcome for the \{side\} side, and devise a new, concise, high-level plan for playing this side in your next game.
Be specific about what you will do differently.
Write in the first person and do not mention player names, since the next game has different players.
\end{quote}

\paragraph{Self-Refine playbook.}
Following Self-Refine~\citep{madaan2023self}, GPT-5-mini writes a strategy playbook for one game from its rules alone, with the same writing instructions as the reflection prompt but no trajectories; it then critiques the draft and revises it, for two rounds in one conversation.
The final text takes the playbook slot.
The playbooks are 1{,}209 (Werewolves), 1{,}273 (Spyfall), and 1{,}334 (Resistance) words long.
Qwen3-32B follows the same procedure, and its playbooks are 1{,}664 (Werewolves), 1{,}444 (Spyfall), and 1{,}562 (Resistance) words long.
GPT-5-mini's Resistance playbook instructs a Spy to fail every mission on which both Spies sit; Spies using it fail 38 missions together, against 15 in vanilla play, and win 1 of 60 games.

\section{Reflection Examples}
\label{sec:appendix:reflections}

Below is an excerpt from GPT-5-mini's $1{\to}1$ Werewolves playbook, which produced the gain reported in \Cref{fig:spartan_delta}.
The full text continues with sections on detection, persuasion, information management, coalition dynamics, and timing.

\begin{quote}
\small\itshape
I want a compact set of habits and rules I can call up during any social deduction game. These are practical, transferable moves --- phrased as things I should do or watch for --- that improve my odds whether I'm town, wolf, or a neutral role.\\
Deception (how I hide or misdirect)\\
- I should build a consistent persona before I start lying. Small consistent details (why I was quiet, what I observed) are cheaper to invent and easier to keep than grand fabrications. Inconsistency is the fastest way to be exposed.\\
- When I need to bluff a role or claim, I should pick claims that are verifiable by game events or that don't force immediate, risky checks. A narrow, time-anchored claim (``I inspected X on night 1; result: villager'') is safer than sweeping claims about many nights.\\
- If I'm going to fabricate an explanation for prior behavior, I should reuse it everywhere: in replies, in votes, in direct answers. Don't invent ad-hoc excuses under pressure.\\
- Avoid overclaiming. The moment I claim a high-value role, I become a target. If I claim, I must be ready to justify my timing and expected consequences (e.g., ``I reveal now because I was targeted last night and needed protection'').\\
- Use silence strategically: being quiet can mask a role, but it invites suspicion.
If I intend to stay quiet, I should state my silence strategy early (``I was listening for tone and alignment'') so the silence itself becomes explainable.
\end{quote}

\section{Game Suite Details}
\label{sec:appendix:games}

\Cref{tab:taxonomy} summarizes the 22 games along the axes used to define our five categories: number of players, player format, information structure, communication mode, iteration horizon, and primary skills exercised.
Full FSM specifications, payoff matrices, and reward functions are released with the code repository alongside the unified game engine.

\begin{table*}[t]
\centering
\resizebox{\textwidth}{!}{
    \begin{tabular}{l l c l l l l l}
    \toprule
    \textbf{Game} & \textbf{Category} & \textbf{Pl.} & \textbf{Format} & \textbf{Information} & \textbf{Comm.} & \textbf{Iteration} & \textbf{Skills} \\
    \midrule
    Prisoner's Dilemma & Normal-Form & 2 & Two-player & Complete & None & 5 rounds & Strategic, Cooperation \\
    Chicken & Normal-Form & 2 & Two-player & Complete & None & 10 rounds & Strategic, Coordination \\
    Battle of the Sexes & Normal-Form & 2 & Two-player & Complete & Free-form & 10 rounds & Strategic, Coordination \\
    Stag Hunt & Normal-Form & 4 & Free-for-all & Complete & None & 10 rounds & Strategic, Cooperation, Coordination \\
    Minority Game & Normal-Form & 5 & Free-for-all & Complete & None & 12 rounds & Strategic, Coordination \\
    Rock-Paper-Scissors & Normal-Form & 2 & Two-player & Complete & None & 10 rounds & Strategic \\
    \midrule
    Public Goods & Economic & 4 & Free-for-all & Hidden state & None & 10 rounds & Strategic, Cooperation \\
    Centipede & Economic & 2 & Two-player & Complete & None & 4 rounds & Strategic, Cooperation \\
    Bargaining & Economic & 2 & Two-player & Complete & None & 10 rounds & Strategic, Negotiation \\
    \href{https://boardgamegeek.com/boardgame/11/bohnanza}{Bohnanza} & Economic & 4 & Free-for-all & Hidden state & Free-form & 6 rounds & Strategic, Negotiation, ToM \\
    \midrule
    \href{https://boardgamegeek.com/boardgame/45/liars-dice}{Liar's Dice} & Bluffing & 3 & Free-for-all & Hidden state & Structured & Until elim. & Deception, Probabilistic \\
    \href{https://boardgamegeek.com/boardgame/92415/skull}{Skull} & Bluffing & 4 & Free-for-all & Hidden state & None & Until 1 round win & Deception, ToM \\
    \href{https://boardgamegeek.com/boardgame/131357/coup}{Coup} & Bluffing & 4 & Free-for-all & Hidden state & Structured & Until elim. & Deception, ToM \\
    \href{https://boardgamegeek.com/boardgame/157969/sheriff-of-nottingham}{Sheriff of Nottingham} & Bluffing & 4 & Free-for-all & Hidden state & Free-form & 4 rounds & Deception, Negotiation, Probabilistic \\
    \midrule
    \href{https://boardgamegeek.com/boardgame/227072/the-chameleon}{Chameleon} & Deduction & 5 & Two-team & Hidden roles & Free-form & 1 round & Deception, Persuasion, ToM \\
    \href{https://boardgamegeek.com/boardgame/206051/insider}{Insider} & Deduction & 5 & Two-team & Hidden roles & Free-form & 1 round & Coordination, Deception, ToM \\
    \href{https://boardgamegeek.com/boardgame/166384/spyfall}{Spyfall} & Deduction & 4 & Two-team & Hidden roles & Free-form & Until elim. & Deception, Persuasion, ToM \\
    \href{https://yanstarstudio.com/undercover-how-to-play}{Undercover} & Deduction & 6 & Two-team & Hidden roles & Free-form & Until elim. & Deception, Persuasion, ToM \\
    \href{https://boardgamegeek.com/boardgame/41114/the-resistance}{Resistance} & Deduction & 5 & Two-team & Hidden roles & Free-form & $\le$5 missions & Strategic, Deception, ToM \\
    \href{https://boardgamegeek.com/boardgame/25821/the-werewolves-of-millers-hollow}{Werewolves} & Deduction & 6 & Two-team & Hidden roles & Free-form & Until elim. & Strategic, Deception, ToM \\
    \midrule
    \href{https://en.wikipedia.org/wiki/Survivor_(American_TV_series)}{Survivor} & Social Strategy & 6 & Free-for-all & Complete & Free-form & Until 3 left & Negotiation, Persuasion, ToM \\
    \href{https://boardgamegeek.com/boardgame/184522/dead-last}{Dead Last} & Social Strategy & 6 & Free-for-all & Complete & Free-form & Until 3 left & Negotiation, Persuasion, ToM \\
    \bottomrule
    \end{tabular}
}
\caption{Game taxonomy across the 22 games in \socialgym.
\textit{Pl.}: number of players; \textit{Format}: player format (\Cref{subsec:game_taxonomy}); \textit{Comm.}: Communication (Format); \textit{ToM}: Theory of Mind.
``Complete'' information means no private state at game start; ``Hidden state'' means each agent has private state (cards, dice) but no factional roles; ``Hidden roles'' means agents are assigned secret allegiances at game start.
\textit{Cooperation} skills involve overcoming the temptation to defect for collective benefit (PD-style); \textit{Coordination} skills involve aligning on one of multiple equilibria (Chicken/BoS-style).
The \textit{Iteration} column uses each game's natural unit: \textit{rounds} (one independent play of the base game; for matrix games each round is one simultaneous move per player, for Centipede each round is one full traversal of the take-or-pass tree), \textit{missions} (Resistance has up to 5 mission proposals), and \textit{Until elim.}
(game ends when a win condition is met, e.g., one team eliminated or last survivor remaining).}
\label{tab:taxonomy}
\end{table*}

\paragraph{Modifications from published versions.}
A few of our commercial-game implementations adopt a slightly modified version of the base game, typically to keep the focus on social reasoning and avoid bookkeeping that the engine would have to track but that does not exercise additional social skills.
We note these here for clarity.

\begin{itemize}
    \item \textbf{\href{https://boardgamegeek.com/boardgame/131357/coup}{Coup}}.
We follow the base-game rules, including the full block matrix, challengeable blocks, the Ambassador's exchange, and the card swap-back after a failed challenge.
    \item \textbf{\href{https://boardgamegeek.com/boardgame/92415/skull}{Skull}}.
A round win ends the game (\texttt{wins\_needed = 1}) rather than 2, shortening the match while keeping the place--bid--flip cycle intact.
    \item \textbf{\href{https://boardgamegeek.com/boardgame/157969/sheriff-of-nottingham}{Sheriff of Nottingham}}.
Goods are summarized as a binary \emph{honest}/\emph{smuggle} choice with fixed payoffs; the inventory layer of the published game (multiple legal goods, royal-goods bonuses) is folded into a single payoff parameter so that the negotiation/inspection dynamic is what drives play.
    \item \textbf{\href{https://boardgamegeek.com/boardgame/206051/insider}{Insider}}.
The environment answers yes/no questions directly, replacing the published \emph{Master} role; the Insider role itself (a Citizen who covertly knows the word and steers questioning) is unchanged.
    \item \textbf{\href{https://boardgamegeek.com/boardgame/166384/spyfall}{Spyfall}}.
The round ends by vote rather than by a real-time clock, and the Spy does not pre-empt the vote with a mid-round location guess.
    \item \textbf{\href{https://boardgamegeek.com/boardgame/11/bohnanza}{Bohnanza}}.
The game ends the first time the draw pile runs out, or after six rounds, rather than the third time, and each player speaks once per trading phase.
    \item \textbf{\href{https://boardgamegeek.com/boardgame/45/liars-dice}{Liar's Dice}}.
Three players hold five dice each, and ones are not wild.
    \item \textbf{\href{https://boardgamegeek.com/boardgame/227072/the-chameleon}{The Chameleon}}.
Each episode is one five-player round with one discussion turn per player and custom 16-word topic cards, without the multi-round point system.
    \item \textbf{\href{https://boardgamegeek.com/boardgame/25821/the-werewolves-of-millers-hollow}{Werewolves}}.
The Seer and Witch follow \emph{The Werewolves of Millers Hollow}.
As in the common Chinese ruleset, the Witch uses at most one potion per night and roles stay hidden on death.
    \item \textbf{\href{https://yanstarstudio.com/undercover-how-to-play}{Undercover}}.
We drop the optional Mr.~White role.
    \item \textbf{\href{https://boardgamegeek.com/boardgame/184522/dead-last}{Dead Last}}.
We keep the discussion and the simultaneous elimination vote.
We drop the penalties for tied targets, for voters outside the plurality, and for Ambush cards, and we require a strict majority, so that players must build a coalition to eliminate anyone.
The game ends with three survivors, who split 100 points only if all three propose the same split, in place of the two-player Share/Steal showdown and the race to 25 points.
    \item \textbf{\href{https://en.wikipedia.org/wiki/Survivor_(American_TV_series)}{Survivor}}.
Eliminations and the jury vote follow the format of the US television show.
\end{itemize}

\noindent The remaining games follow their canonical or published mechanics: Resistance, Bargaining (iterated ultimatum), and the textbook normal-form games (PD, Chicken, Stag Hunt, Battle of the Sexes, Centipede, Public Goods, Minority Game, Rock-Paper-Scissors).
Setup choices such as fixed player counts, the specific role roster in Werewolves, the absence of ``wild ones'' in Liar's Dice, and two Undercovers without \emph{Mr.\ White} sit within the configuration space of the published games.

\section{Behavioral Measures and Rating Checks}
\label{sec:appendix:behaviors}

This appendix gives the measures and further results behind \Cref{sec:results:secrets}, grouped by its three questions, followed by the random-player comparison, the discussion experiment, the rating checks, and the outside benchmarks.
For the behavioral measures, we replay recorded actions through the engine version used for each episode.
This recovers the votes, team proposals, bribe offers, and trades that the engine accepted during play.
Chance rates use uniformly random choices from the legal options for the corresponding action.
Each subsection pairs a measure's definition with its results and its per-model table.

\subsection{Keeping information private}
\label{sec:appendix:beh_private}

\paragraph{Chameleon discussion disclosure.}
We count Chameleon episodes in which a Citizen states the secret word in public discussion (exact match, case-insensitive), out of all of a model's Citizen episodes, as in \Cref{fig:behaviors}.
To separate discussion leaks from clue leaks, we also restrict the count to episodes in which no player, not even the Chameleon, gave the secret word as a clue.
In these episodes, the Citizens of GPT-5-mini and GPT-4o name the word during discussion in 68\% and 96\% of games.
The Chameleon guess rates in \Cref{sec:results:secrets} use these episodes without a clue leak.
A manual check of 40 sampled episodes found 36 that state or confirm the word, and 4 that name it only as one option.

\paragraph{Spyfall disclosure.}
We count a Spyfall episode as a disclosure when a Non-Spy states the location name in public speech (case-insensitive match, allowing plurals and hyphenation).
In 15 of the 91 such episodes the Spy names the location first and a Non-Spy repeats it.
Counting only episodes in which a Non-Spy names it first, the six models (GPT-4o-mini, Gemma3-4B, and the four Qwen models) that disclose the location in at least 19\% of their games range from 11 to 78\%.
A manual check of 32 sampled disclosures confirmed 30.

\subsection{Acting on evidence}
\label{sec:appendix:beh_evidence}

\paragraph{Werewolves votes.}
We analyze 2{,}082 day votes from 270 episodes.
A village vote is correct if it names a living Werewolf.
Votes on dead players count as incorrect.
\Cref{tab:appx_ww} gives the per-model results.

\begin{table}[h]
\centering
\footnotesize
\resizebox{\linewidth}{!}{
\begin{tabular}{lrrrrrrr}
\toprule
Model & \shortstack{Village\\votes} & \shortstack{On a\\Werewolf (\%)} & \shortstack{Chance\\(\%)} & \shortstack{Day 1\\(\%, $n$)} & \shortstack{Day 2\\(\%, $n$)} & \shortstack{Werewolf\\votes} & \shortstack{On partner\\(\%)} \\
\midrule
GPT-5-mini & 151 & 77 & 39 & 73 (108) & 86 (43) & 58 & 0 \\
GPT-4o & 149 & 76 & 41 & 72 (108) & 85 (39) & 65 & 3 \\
GPT-4o-mini & 149 & 67 & 42 & 67 (100) & 68 (47) & 62 & 8 \\
Gemma3-27B & 147 & 65 & 41 & 60 (108) & 74 (34) & 65 & 8 \\
Qwen3-32B & 137 & 63 & 40 & 59 (106) & 74 (31) & 66 & 9 \\
Gemma3-12B & 160 & 59 & 40 & 55 (100) & 62 (45) & 62 & 13 \\
Gemma3-4B & 137 & 36 & 42 & 42 (100) & 16 (25) & 74 & 23 \\
Qwen3-4B & 119 & 37 & 43 & 45 (91) & 14 (21) & 74 & 1 \\
Qwen2.5-3B & 120 & 35 & 43 & 38 (82) & 33 (24) & 85 & 11 \\
Qwen3-1.7B & 125 & 16 & 43 & 20 (84) & 5 (19) & 74 & 36 \\
\bottomrule
\end{tabular}
}
\caption{Werewolves day votes per model.
Village votes are those of the Villager, Seer, and Witch; a vote is on a Werewolf when it names a living Werewolf.
Chance is the share a uniformly random vote among the other living players would put on a Werewolf.
The last two columns give Werewolves' votes on their living partner.}
\label{tab:appx_ww}
\end{table}

\paragraph{Votes across hidden-role games.}
We apply the Werewolves measure to all six hidden-role games.
A vote counts when a majority-side player votes for a hidden opponent still in the game, and chance is the share a uniform vote among the voter's eligible targets would put there.
Resistance has no vote that names a player, so we use whether a Resistance leader's proposed team holds no Spy, against a uniformly drawn team of the same size.
A model shows no reliable advantage over chance in a game when the 95\% Wilson interval of its rate reaches the chance rate (\Cref{tab:appx_votes}).
This does not show that the model plays at chance.
Two small models stand out in single games.
Qwen3-1.7B finds the Insider in 75\% of its Insider votes (\Cref{sec:results:secrets}), and Gemma3-4B votes for the Spy in 60\% of its Spyfall votes.

\begin{table}[h]
\centering
\small
\resizebox{\linewidth}{!}{
\begin{tabular}{lrrrrrr}
\toprule
Model & Werewolves & Undercover & Spyfall & Chameleon & Insider & Resistance \\
\midrule
GPT-5-mini & 77 & 99 & 60 & 82 & 27$^\dagger$ & 57 \\
GPT-4o & 76 & 88 & 64 & 57 & 48 & 54 \\
GPT-4o-mini & 67 & 84 & 44$^\dagger$ & 29$^\dagger$ & 20$^\dagger$ & 47 \\
Gemma3-27B & 65 & 75 & 51 & 37 & 19$^\dagger$ & 20$^\dagger$ \\
Qwen3-32B & 63 & 72 & 51 & 24$^\dagger$ & 48 & 20$^\dagger$ \\
Gemma3-12B & 59 & 68 & 61 & 32$^\dagger$ & 20$^\dagger$ & 19$^\dagger$ \\
Gemma3-4B & 36$^\dagger$ & 40$^\dagger$ & 60 & 6$^\dagger$ & 44 & 31$^\dagger$ \\
Qwen3-4B & 37$^\dagger$ & 55 & 41$^\dagger$ & 24$^\dagger$ & 23$^\dagger$ & 29 \\
Qwen2.5-3B & 35$^\dagger$ & 46$^\dagger$ & 32$^\dagger$ & 28$^\dagger$ & 19$^\dagger$ & 22$^\dagger$ \\
Qwen3-1.7B & 16$^\dagger$ & 38$^\dagger$ & 31$^\dagger$ & 20$^\dagger$ & 75 & 19$^\dagger$ \\
\midrule
Chance & 39--43 & 33--44 & 36--39 & 25 & 25 & 18--26 \\
\bottomrule
\end{tabular}
}
\caption{Majority-side votes on a hidden opponent (\%), per model and hidden-role game.
Resistance: share of a Resistance leader's proposed teams with no Spy.
Chance: a uniform vote (or team) among the eligible options, which varies with the number of living players.
$^\dagger$No reliable advantage over chance: the 95\% Wilson interval reaches the chance rate.}
\label{tab:appx_votes}
\end{table}

\paragraph{Resistance teams and votes.}
We analyze 1{,}265 proposed teams, 3{,}795 team votes by Resistance players, and 965 missions from 270 episodes.
For each proposed team, we record whether it contains a Spy.
We then compare Resistance players' approval rates for teams with and without a Spy.
\Cref{tab:appx_rs} gives the per-model results.

\begin{table}[h]
\centering
\footnotesize
\resizebox{\linewidth}{!}{
\begin{tabular}{lrrrrrr}
\toprule
Model & \shortstack{Teams led\\as Res.} & \shortstack{With a\\Spy (\%)} & \shortstack{Teams led\\as Spy} & \shortstack{With a\\Spy (\%)} & \shortstack{Approve Spy\\teams (\%)} & \shortstack{Approve clean\\teams (\%)} \\
\midrule
GPT-5-mini & 82 & 43 & 53 & 100 & 73 & 96 \\
GPT-4o & 74 & 46 & 49 & 92 & 82 & 93 \\
GPT-4o-mini & 79 & 53 & 65 & 92 & 57 & 74 \\
Gemma3-27B & 74 & 80 & 46 & 74 & 86 & 89 \\
Qwen3-32B & 65 & 80 & 40 & 60 & 97 & 99 \\
Gemma3-12B & 73 & 81 & 49 & 63 & 91 & 99 \\
Gemma3-4B & 59 & 69 & 50 & 44 & 99 & 78 \\
Qwen3-4B & 75 & 71 & 54 & 57 & 72 & 82 \\
Qwen2.5-3B & 102 & 78 & 55 & 73 & 22 & 7 \\
Qwen3-1.7B & 75 & 81 & 46 & 48 & 77 & 76 \\
\bottomrule
\end{tabular}
}
\caption{Resistance team proposals and votes per model.
The first four data columns report the number of team proposals made as a Resistance or Spy leader and the percentage containing at least one Spy (a random team contains one 79\% of the time).
Each proposal counts once, whether or not the team is approved.
The last two columns give the share of proposed teams the model approved as a Resistance voter, split by whether the team contained a Spy.}
\label{tab:appx_rs}
\end{table}

\paragraph{Alliances and joint vote calls.}
We analyze 4{,}769 speeches from 90 Survivor and 150 Dead Last episodes.
Qwen3.8-27B labels declarations of an alliance with a named player and calls for a joint vote on a named target.
The judge reads each speech with the game rules, phase, and player list, without the full conversation history.
We check whether the speaker's next vote spares the declared ally or targets the player named in the call.
The kept rates do not depend on the labeler: Qwen3.8-27B, Claude Haiku 5.5, and a rule-based labeler that matches fixed sentence patterns give kept rates within 2 points of each other and nearly the same order of models (Spearman 0.88).
Because Qwen3.8-27B finds labels that the rule-based labeler misses, we manually checked 40 such labels from 30 speeches and confirmed 34.

\paragraph{Following and leading votes.}
In Survivor and Dead Last, a model that calls on the other players to vote someone out usually votes that way itself: its next vote follows the call 83\% of the time, against 25\% by chance.
Weak models break their calls by voting for players nobody named, while strong models switch to another player's call once their own loses support.
The other players often follow too: after a call, 65\% of their votes go to the named player (chance 24\%).
GPT-5-mini is the most persuasive: after its first call on a player, 74\% of the other players' votes go to that player, who goes out 69\% of the time (\Cref{tab:appx_al}).

\begin{table}[h]
\centering
\footnotesize
\resizebox{\linewidth}{!}{
\begin{tabular}{lrrrrrr}
\toprule
Model & Speeches & \shortstack{With a\\declaration (\%)} & \shortstack{Ally\\checks} & \shortstack{Kept\\(\%)} & \shortstack{Joint vote\\checks} & \shortstack{Kept\\(\%)} \\
\midrule
GPT-5-mini & 527 & 70 & 322 & 98 & 319 & 92 \\
GPT-4o & 474 & 23 & 23 & 83 & 93 & 86 \\
GPT-4o-mini & 462 & 39 & 21 & 90 & 170 & 81 \\
Gemma3-27B & 493 & 58 & 75 & 92 & 257 & 89 \\
Qwen3-32B & 469 & 28 & 80 & 91 & 106 & 88 \\
Gemma3-12B & 513 & 42 & 74 & 91 & 170 & 88 \\
Gemma3-4B & 381 & 44 & 34 & 79 & 137 & 59 \\
Qwen3-4B & 491 & 68 & 103 & 85 & 289 & 87 \\
Qwen2.5-3B & 469 & 50 & 46 & 83 & 175 & 62 \\
Qwen3-1.7B & 490 & 44 & 18 & 83 & 182 & 82 \\
\bottomrule
\end{tabular}
}
\caption{Alliance declarations in Survivor and Dead Last per model, labelled by Qwen3.8-27B.
An ally declaration is kept when the speaker's next vote spares the named player (73\% by chance); a joint vote call is kept when the next vote lands on the called target (25\% by chance).}
\label{tab:appx_al}
\end{table}

\paragraph{Blending in.}
In Undercover, the minority wins by sounding like the majority.
Because nobody is told their side, an Undercover must first work out from the others' descriptions that its word is the minority one.
Strong Civilians catch Undercovers quickly, putting 90 to 98\% of their first-round votes on one (chance 40\%), and no model is strong as an Undercover (ratings 781 to 995).
Only GPT-5-mini, as an Undercover, switches to describing the majority word after hearing the others.
In one game, the GPT-4o Civilians hold \emph{telescope} and two GPT-5-mini Undercovers hold \emph{microscope}.
After a Civilian says ``You might find this at an observatory,'' the first Undercover says ``I would bring this to an observatory to study distant objects in the sky,'' and the Undercovers go on to win.
Undercovers who blend in this way go out in the first round 19\% of the time, against 37\% for those who keep describing their own word.

\paragraph{Winning without good questions.}
Citizens who ask weak questions can still win.
In Insider, everyone loses if the group never finds the word, so the Insider must help the group without standing out.
Qwen3-1.7B's Citizens ask questions that rarely get a Yes (16\%, against 39 to 71\% for the other models), so the Insider has to name the word itself and is the only correct guesser in 80\% of these games.
One Citizen's accusation, copied word for word by the others, then lands on the Insider, and the Qwen3-1.7B Citizens catch the Insider in 76\% of the games where the word is found.
GPT-5-mini's Citizens find the word together, so a correct guess reveals nothing, and they catch the Insider in only 26\%.

\subsection{Serving the game's goal}
\label{sec:appendix:beh_goal}

\paragraph{Sheriff bribes and inspections.}
We analyze 1{,}620 merchant bags from 135 episodes.
A bag is counted as bribed when the engine records a positive bribe offer in the merchant's statement.
We record whether each bag contains contraband and whether the Sheriff inspects it.
If the Sheriff passes a bribed bag, the merchant pays the bribe, up to the gold it has.
Sheriffs as a group do not treat a bribe as a sign of contraband: they inspect bribed bags about as often as the others (53 against 55\%), although bribed bags carry contraband far more often (85 against 29\%).
The two models that play Sheriff best take the opposite approach: Qwen3-1.7B and Gemma3-4B rarely smuggle as merchants (2\% and 6\% of bags) and inspect most bribed bags as Sheriffs (84\% and 64\%), ranking first and second in Sheriff, nearly opposite to their leaderboard positions (\Cref{tab:appx_sh}).

\paragraph{Bargaining offers.}
In Bargaining, two players take turns proposing how to split 10 tokens over 10 rounds, and a rejected split gives both players nothing; we count an offer as below half when it leaves the responder fewer than 5 tokens (270 episodes, 54 per model).
Because the player with more tokens wins, accepting a below-half offer adds a few tokens but gives the proposer more, whereas rejecting it leaves the gap unchanged.
Across the ten models, the more below-half offers a model accepts, the fewer games it wins (Spearman $-0.57$).
For example, GPT-5-mini accepts 81\% of such offers and wins 56\% of its games, while GPT-4o accepts 22\% and wins 76\%.

\begin{table*}[h]
\centering
\footnotesize
\resizebox{\textwidth}{!}{
\begin{tabular}{lrrrrrrrr}
\toprule
Model & \shortstack{Bribes (\% of\\statements)} & \shortstack{Mean\\offer} & \shortstack{Smuggle\\(\%)} & \shortstack{Smuggle,\\bribed (\%)} & \shortstack{Smuggle,\\no bribe (\%)} & \shortstack{Inspect\\bribed (\%)} & \shortstack{Inspect\\other (\%)} & \shortstack{Bribe gold\\per round} \\
\midrule
GPT-5-mini & 64 & 4.44 & 62 & 97 & 0 & 20 & 36 & 6.39 \\
GPT-4o & 88 & 2.96 & 70 & 76 & 26 & 29 & 69 & 4.26 \\
GPT-4o-mini & 81 & 3.11 & 83 & 81 & 90 & 54 & 47 & 3.63 \\
Gemma3-27B & 94 & 4.45 & 92 & 92 & 90 & 34 & 46 & 4.67 \\
Qwen3-32B & 75 & 4.81 & 98 & 98 & 95 & 59 & 65 & 2.83 \\
Gemma3-12B & 94 & 4.67 & 93 & 97 & 30 & 37 & 55 & 4.50 \\
Gemma3-4B & 16 & 4.50 & 6 & 19 & 3 & 64 & 40 & 3.15 \\
Qwen3-4B & 90 & 5.30 & 98 & 99 & 94 & 53 & 53 & 4.13 \\
Qwen2.5-3B & 52 & 3.81 & 63 & 60 & 67 & 82 & 70 & 1.39 \\
Qwen3-1.7B & 14 & 3.64 & 2 & 0 & 2 & 84 & 79 & 1.57 \\
\bottomrule
\end{tabular}
}
\caption{Sheriff per model: as a merchant (statements that bind a bribe, mean offer in gold, share of bags with contraband, overall and split by whether the merchant offered a bribe) and as Sheriff (inspection rate of bribed and other bags, bribe gold collected per round).}
\label{tab:appx_sh}
\end{table*}

\paragraph{Bohnanza trades.}
We analyze 12{,}889 offers and 4{,}691 completed trades from 135 episodes; 4{,}289 of the trades are swaps, in which both players give cards, and the other 402 are gifts.
We follow each received card through the subsequent planting phase and count a trade as helping a player when a card it receives is planted on a field already growing that variety.
Models trade often but gain little from it.
Of the 4{,}289 swaps, 76\% help neither player, because 79\% of received cards are a variety for which the receiver has no field.
The receiver must then uproot a field to plant them.
Across all players, the number of trades a player completes is unrelated to its final coins (Spearman 0.05).
GPT-5-mini is the exception: of the cards it receives, 36\% extend one of its fields, against 13\% across all players and 5\% for GPT-4o, and it ends with the most coins (\Cref{tab:appx_bo}).

\begin{table*}[h]
\centering
\footnotesize
\resizebox{\textwidth}{!}{
\begin{tabular}{lrrrrrrr}
\toprule
Model & \shortstack{Offers\\per game} & \shortstack{Completed\\(\%)} & \shortstack{Accepted of\\received (\%)} & \shortstack{Trades\\per game} & \shortstack{Cards matching\\a field (\%)} & \shortstack{Cards extending\\a field (\%)} & \shortstack{Mean\\coins} \\
\midrule
GPT-5-mini & 31.70 & 37 & 26 & 17.11 & 37 & 36 & 2.07 \\
GPT-4o & 30.94 & 39 & 26 & 17.63 & 19 & 5 & 0.52 \\
GPT-4o-mini & 26.33 & 43 & 39 & 19.93 & 19 & 10 & 0.65 \\
Gemma3-27B & 29.69 & 31 & 42 & 18.43 & 19 & 16 & 1.22 \\
Qwen3-32B & 24.26 & 35 & 36 & 17.15 & 23 & 15 & 0.93 \\
Gemma3-12B & 22.48 & 30 & 39 & 16.33 & 20 & 11 & 0.80 \\
Gemma3-4B & 17.07 & 39 & 45 & 18.00 & 18 & 11 & 0.85 \\
Qwen3-4B & 24.24 & 40 & 33 & 17.85 & 17 & 5 & 0.46 \\
Qwen2.5-3B & 17.76 & 43 & 29 & 15.67 & 24 & 11 & 0.74 \\
Qwen3-1.7B & 14.20 & 23 & 47 & 15.65 & 18 & 11 & 0.63 \\
\bottomrule
\end{tabular}
}
\caption{Bohnanza trading per model: offers posted per player-game and the share completed, the share of received offers the model accepted, trades joined per game, the share of received cards that matched a field the model held and that were planted on it, and final coins.}
\label{tab:appx_bo}
\end{table*}

\subsection{Random-player comparison}
\label{sec:appendix:random}

\begin{figure}[t]
\centering
\includegraphics[width=\linewidth]{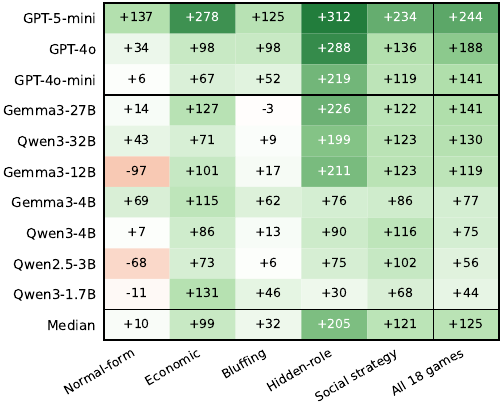}
\caption{Elo above a uniformly random player, by model and game family (each family's scale shifted so that the random player sits at 0; green: above random).}
\label{fig:elo_above_random}
\end{figure}

To give the ratings an absolute reference point, we add a scripted player that selects uniformly from the legal options.
When an action has many variants, it first selects an action type and then a variant.
In two-player and team games, the random player fills the opposing side, with 15 episodes per model and game.
In the five free-for-all games, it takes one seat at a table of models, with 90 episodes per game.
Using one random seat retains the other models' votes and negotiation, which would also become random if every opponent were replaced.
We fit ratings jointly on these episodes and the leaderboard episodes, fixing the random player's Elo at 1000.

Over the competitive games, every model rates above the random player, but the gap varies across game families: the median model is 205 Elo above random in the hidden-role games, 99 in the economic games, and 10 in the normal-form games (\Cref{fig:elo_above_random}).
The economic median is lower than the 262 of Bargaining alone because four models play Bohnanza about as well as random.
The normal-form gap is small partly by design: in Rock-Paper-Scissors and the Minority Game, random play is an equilibrium, so no strategy beats it by much.
Two models fall clearly below random in the normal-form games, Gemma3-12B ($-97$) and Qwen2.5-3B ($-68$).

\subsection{Discussion experiment}
\label{sec:results:talk}

\begin{figure}[t]
\centering
\includegraphics[width=\linewidth]{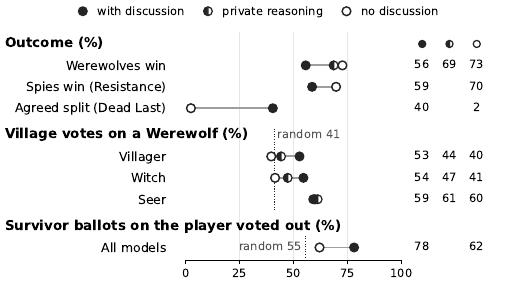}
\caption{Play with discussion, with a private reasoning turn in its place (Werewolves only), and with no discussion, on the same deals.
Agreed split: Dead Last finals in which all three finalists propose the same split.
Survivor ballots: share of ballots that name the player voted out that round; its random line is simulated, because the player voted out is the one with the most votes.}
\label{fig:comm_control}
\end{figure}

To test how much of what \socialgym measures in dialogue games depends on what players say, we replay four of them without their discussion phase: Werewolves, Resistance, Survivor, and Dead Last (\Cref{fig:comm_control}).
These four multi-party dialogue games remain playable after we remove the discussion phase.
In the other six multi-party dialogue games, dialogue is itself a game action, such as the clues in Chameleon.
Each game is played in both conditions on the same model assignments, seats, roles, and initial deals, and every difference is paired by deal (270 Werewolves, 270 Resistance, 90 Survivor, and 150 Dead Last episodes per condition).
Removing the discussion phase also removes it from the rules text, while votes, proposals, mission results, and every other action and win condition stay the same.
Survivor has no sides, and its winner changes in 73\% of deals even between two plays with discussion, so for Survivor \Cref{fig:comm_control} shows votes rather than win rates.

In Werewolves, removing discussion keeps the Seer from sharing its inspection results: the Seer itself still votes for a Werewolf 60\% of the time, but the other Villagers' accuracy falls from 53\% to 40\%, close to a random vote (41\%).
On the first day, an ordinary Villager then sees only who died and has no evidence about the Werewolves.
As a result, the Werewolves' win rate rises from 56\% to 73\%, and GPT-5-mini's Villagers win 30\% of games instead of 96\%.
Resistance changes less, because players still see team proposals, votes, and mission results without discussion: the Spies' win rate rises from 59\% to 70\%.
In Survivor, the share of ballots on the player voted out falls from 78\% to 62\%, close to the 55\% that random voting gives.
Dead Last changes most, because without discussion the three finalists must propose the same split without coordinating: the share of finals ending in an agreed split falls from 40\% to 2\%, and every model's win rate falls to 2\% or below.
Discussion is thus where the models differ most: without it, the Werewolves ratings keep nearly the leaderboard's order (Spearman 0.91), but their range shrinks from 725 to 232 Elo.

Removing discussion also removes the reasoning a model writes out before it votes, so the drop could come from either.
A third condition separates the two by replacing each discussion turn with a private reasoning turn that no other player sees.
Private reasoning recovers little: the Werewolves win 69\% of deals, close to the 73\% without discussion, and GPT-5-mini's Villagers win 22\%, far below the 96\% they win with discussion.
The gain from discussion therefore comes from what the other players say.

\subsection{Rating checks}

\paragraph{Ranking stability.}\label{sec:appendix:stability}
The overall ranking holds on much less data.
Each check fits ratings on part of the tournament and compares the resulting ranking with the full one.
A random fifth of the episodes (800) already reproduces the overall ranking at Spearman 0.94.
The ranking is also stable across deal order and seat position.
Ratings fit on the earlier and the later half of the deals agree at 0.88.
Ratings fit on early and late seats agree at 0.93.
GPT-5-mini and GPT-4o rank first and second in every half.
Single games see only 54 episodes per model, yet their rankings are also stable under bootstrap resampling.
The median Spearman between a resampled and the full-data ranking is at least 0.90 for 17 of the 18 games, and 0.66 for Dead Last.

\paragraph{Splitting the evaluation data.}
For the deal-order check, we split episodes into earlier and later halves within each roster set, model pair, and side assignment.
Episodes seating three or more different models are split within their roster set.
For bootstrap rank stability, we compare the ranking in each of the 1{,}000 pair-bootstrap draws behind the rating CIs with the full-data ranking and report the median Spearman correlation.
For the seat-order check, we fit separate ratings for each model in early and late seats.
Seats in the first half of the player list are early and the remainder late; for an odd number of players, the middle seat is early.

\paragraph{Game-family rankings.}
\Cref{tab:subset} compares each game-family ranking with the overall ranking and reports its split-half reliability.
The normal-form, bluffing, and economic rankings correlate with the overall ranking far below the ceiling their reliability allows (about 0.94, 0.95, and 0.82), so their differences from the overall ranking are not noise.
An equal-weight average of the five family ratings reproduces the overall ranking at Spearman 0.96, although the hidden-role games supply 43\% of the episodes.

\begin{table}[h]
\centering
\small
\resizebox{\linewidth}{!}{
\begin{tabular}{lrrrr}
\toprule
Game family & Games & Eps. & \shortstack{$\rho$ with\\overall} & \shortstack{Split-half\\reliability} \\
\midrule
Normal-form & 4 & 24\% & 0.55 & 0.89 \\
Economic & 2 & 11\% & 0.04 & 0.68 \\
Bluffing & 4 & 16\% & 0.47 & 0.90 \\
Hidden-role & 6 & 43\% & 0.96 & 0.94 \\
Social strategy & 2 & 6\% & 0.85 & 0.55 \\
\midrule
All competitive & 18 & 100\% & 1.00 & 0.97 \\
\bottomrule
\end{tabular}
}
\caption{Elo fit on each game family alone.
Games: competitive games in the family.
Eps.: share of episodes.
$\rho$ with overall: Spearman correlation between the family's ranking and the overall ranking.
Split-half reliability: median Spearman correlation between rankings fit on two random halves of the family's episodes, corrected to the full data with the Spearman--Brown formula.
If a family's ranking were the overall ranking plus noise, its $\rho$ with overall would be close to the square root of its reliability.}
\label{tab:subset}
\end{table}

\paragraph{Opponent dependence.}
Elo assumes that a model's chance of beating an opponent depends only on their rating difference, so we check whether a model's results depend on which opponents it faces.
Each model contributes approximately 3{,}700 pairwise score comparisons across the 18 competitive games, using the outcome extraction in \Cref{subsec:elo}; these are comparisons between agents, not separate episodes.
A tie counts as half a win.
For the observed-versus-predicted win-rate check, an opponent is stronger if its overall Elo is more than 30 points higher, similarly rated if it is within 30 points, and weaker if it is more than 30 points lower.
Separately, we refit each model's rating against the top three, middle four, and bottom three models in the leaderboard, keeping opponents' ratings fixed and excluding same-model comparisons.
The observed win rate is within 2 percentage points of the Elo prediction in 21 of 25 (model, opponent group) pairs, and fitting ratings separately against each group changes a model's rating by 8 Elo on average (\Cref{tab:appx_od}).
A model's rating therefore depends little on which opponents it happens to face.

\paragraph{Execution errors.}
All 4{,}588 episodes are rated; nine ended at a step cap.
Malformed moves remain visible to the other players and are handled according to the game rules (\Cref{sec:appendix:models}).
Of the 155{,}971 model calls in the rated episodes, 94.9\% are used as written.
The share that the engine cannot use as written ranges from 0.6\% for GPT-5-mini to 11.1\% for Qwen3-1.7B.
Bohnanza accounts for most of these calls, and without it the range is 0.05\% to 5.9\%.
Because the engine applies each game's own consequence to these moves, such as a lost die or an abstained vote, they count against the model that made them, and they are most common among the smallest models.

\begin{table}[h]
\centering
\footnotesize
\resizebox{\linewidth}{!}{
\begin{tabular}{lrrr}
\toprule
Model & \shortstack{vs.\ stronger:\\win / expected (\%, $n$)} & vs.\ similar & vs.\ weaker \\
\midrule
GPT-5-mini & -- & -- & 68 / 68 (3702) \\
GPT-4o & 38 / 42 (405) & -- & 63 / 62 (3297) \\
GPT-4o-mini & 42 / 40 (835) & 52 / 52 (1267) & 59 / 61 (1600) \\
Gemma3-27B & 39 / 39 (845) & 52 / 51 (1270) & 61 / 61 (1587) \\
Qwen3-32B & 38 / 38 (857) & 49 / 49 (1265) & 60 / 60 (1580) \\
Gemma3-12B & 36 / 36 (807) & 46 / 47 (1088) & 59 / 58 (1807) \\
Gemma3-4B & 38 / 38 (2409) & 54 / 52 (779) & 51 / 55 (514) \\
Qwen3-4B & 38 / 38 (2500) & 50 / 51 (813) & 55 / 55 (389) \\
Qwen2.5-3B & 36 / 36 (2505) & 48 / 49 (1197) & -- \\
Qwen3-1.7B & 37 / 37 (3313) & 49 / 48 (389) & -- \\
\bottomrule
\end{tabular}
}
\caption{Opponent dependence: each model's win share against opponents more than 30 Elo above it, within 30 Elo, and more than 30 Elo below, next to the share its Elo predicts; $n$ is the number of battles.}
\label{tab:appx_od}
\end{table}

\subsection{Outside benchmarks}
\label{sec:appendix:outside}

\begin{figure}[h]
\centering
\includegraphics[width=\linewidth]{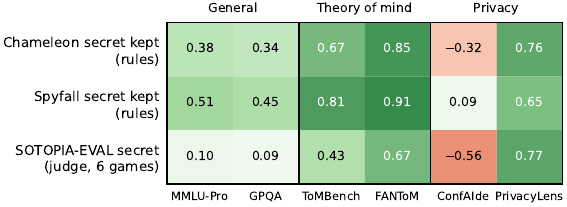}
\caption{Spearman correlation across the ten models between three secret-keeping measures and outside benchmarks (green positive).}
\label{fig:corr_secret}
\end{figure}

\paragraph{Secret keeping and outside benchmarks.}
\Cref{fig:corr_secret} correlates three secret-keeping measures with the outside benchmarks of \Cref{tab:corr}: the share of Chameleon games in which a model, as a Citizen, does not state the secret word as its clue (rules), the share of Spyfall games in which a model's Non-Spies do not name the location in public speech (rules), and the SOTOPIA-EVAL secret dimension scored by Qwen3.8-27B over all six hidden-role games (judge).
All three correlate more strongly with FANToM and PrivacyLens than with the general benchmarks.
ConfAIde is the exception ($-0.32$, $0.09$, and $-0.56$), because its Tier 3 metric counts a reply as a leak whenever it names the person whose secret is at stake.
Gemma3-27B and Gemma3-12B, which keep secrets well in our games, name that person outside the reply they propose in 81\% and 72\% of their answers, when introducing the reply or explaining why it protects that person.
Scoring only the reply a model writes for the speaking character, the ConfAIde correlations become positive (0.37, 0.73, and 0.19).

\begin{figure*}[h]
\centering
\includegraphics[width=0.78\textwidth]{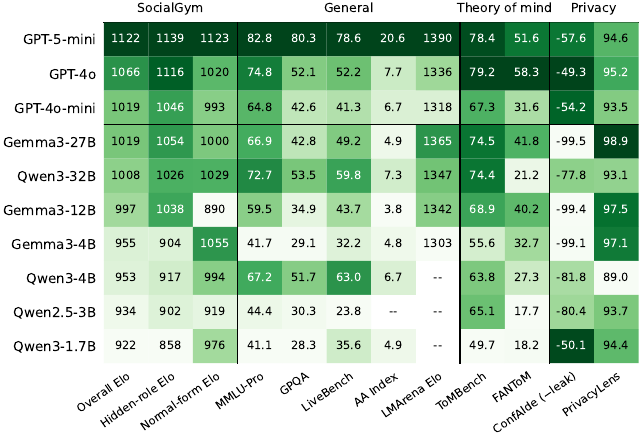}
\caption{Per-model scores on \socialgym measures and outside benchmarks, with each column shaded linearly from its lowest (white) to its highest value (dark green).
Elo values are the leaderboard's overall fit and the hidden-role and normal-form family fits.
MMLU-Pro, GPQA Diamond, LiveBench, the Artificial Analysis Intelligence Index, and the LMArena text Elo are published scores, except that we ran LiveBench for GPT-5-mini and Gemma3-4B and MMLU-Pro and GPQA for Qwen2.5-3B with the published prompts (missing entries were not published); ToMBench, FANToM (belief multiple choice), ConfAIde (Tier 3 leak rate, negated) and PrivacyLens (probing accuracy) were run by us.}
\label{fig:benchmark_scores}
\end{figure*}

\paragraph{Outside benchmark scores.}
\Cref{fig:benchmark_scores} lists every score behind the correlations in \Cref{tab:corr}, with three further general benchmarks (LiveBench, the Artificial Analysis Intelligence Index, and the LMArena text Elo) that cover only part of the model set.

\section{Role-Conditioned Performance and Game Balance}
\label{sec:appendix:balance}

This appendix gives the per-side results behind \Cref{sec:results:sides} for all six hidden-role games.
We first present the per-model cross-play role gap (\Cref{fig:secrets_sides}), then same-model self-play (\Cref{tab:balance_selfplay}), in which one model fills every seat, so that a side's win rate no longer depends on the tournament opponents.

\subsection{Cross-play role gaps}
For each hidden-role game we estimate a separate Elo for the minority/deceptive role (\textbf{Elo-Alt}: Werewolf, Spy, Insider, Chameleon, Undercover) and for the majority/cooperative role (\textbf{Elo-Main}: Villager, Civilian, Non-Spy).
The gaps in \Cref{fig:secrets_sides} are large: $|\Delta|$ exceeds 100 Elo in 37 of the 60 (model, game) cells, with a median of 190.
Five games point the same way for most models: the minority side is the stronger one in Chameleon (all ten models), Insider (eight), and Werewolves (eight), and the weaker one in Spyfall (all ten) and Undercover (eight).
Resistance depends on the model most, ranging from $-220$ (GPT-5-mini) to $+447$ (Qwen2.5-3B).

\begin{figure}[h]
\centering
\includegraphics[width=\linewidth]{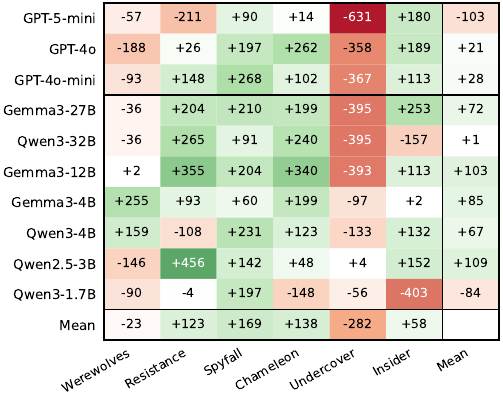}
\caption{Role gap after subtracting the game's own asymmetry: each cell of \Cref{fig:secrets_sides} minus the minority-side edge under all-random play in that game, so that a cell shows how much a model shifts the balance beyond the rules.
Chameleon's edge shrinks and turns negative for Qwen3-1.7B, Spyfall's uniform handicap turns into a uniform gain (the models narrow the Spy's structural disadvantage), and Resistance is unchanged.}
\label{fig:role_gap_corrected}
\end{figure}

To separate the game's own asymmetry from the models', we compare each gap with the gap under all-random play (\Cref{fig:secrets_sides}, Random play row), and \Cref{fig:role_gap_corrected} subtracts this random-play edge from every cell.
Resistance is balanced under random play, so the Spies' advantage, and its reversal for GPT-5-mini, comes from the models.
In Chameleon, every model rates higher on the Chameleon side, by 52 to 540 Elo; random play already gives that side 200 Elo, so the models add 138 Elo on average beyond the game's own edge.
Undercover runs the other way: beyond what random play predicts, the models play the majority side 282 Elo better on average, and the gap grows with model strength, from within 70 Elo of zero for the four smallest models to $-568$ for GPT-5-mini (raw gaps in \Cref{fig:secrets_sides}).
The gap comes from the Civilian side: Civilian ratings range from 815 to 1521 across the models, while Undercover ratings stay within 781 to 995, as strong Civilians put 90 to 98\% of their first-round votes on an Undercover (\Cref{sec:appendix:beh_evidence}).
Same-model self-play provides a complementary comparison with model identity held constant across sides (\Cref{sec:appendix:balance:selfplay}).

\subsection{Same-model self-play}
\label{sec:appendix:balance:selfplay}

Cross-play measures each side against the tournament's model pool, while same-model self-play measures which side a model wins more often when it plays against itself.
A self-play win rate therefore still reflects both the game's own balance and how well the model plays each side.
We report self-play on five hidden-role games in \Cref{tab:balance_selfplay}, with 30 or 60 deals per model and game.

\begin{table}[h]
\centering
\small
\resizebox{\linewidth}{!}{
    \begin{tabular}{l|rrrrr}
    \toprule
    \textbf{Model} & \textbf{Werew.} & \textbf{Spyfall} & \textbf{Cham.} & \textbf{Underc.} & \textbf{Resist.} \\
    \midrule
GPT-5-mini & 37 & 33 & 100 & 7 & 33 \\
GPT-4o & 40 & 23 & 100 & 0 & 67 \\
GPT-4o-mini & 77 & 47 & 97 & 7 & 68 \\
Gemma3-27B & 50 & 23 & 93 & 17 & 93 \\
Qwen3-32B & 67 & 20 & 92 & 13 & 75 \\
Gemma3-12B & 90 & 37 & 93 & 18 & 88 \\
Gemma3-4B & 77 & 42 & 92 & 67 & 45 \\
Qwen3-4B & 73 & 32 & 100 & 45 & 48 \\
Qwen2.5-3B & 57 & 62 & 90 & 53 & 100 \\
Qwen3-1.7B & 58 & 30 & 93 & 57 & 15 \\
    \bottomrule
    \end{tabular}
}
\caption{Same-model self-play: minority-side win rate (\%) per model and game.
Every cell uses 60 deals, except that GPT-4o and GPT-4o-mini use 30 in all games but Resistance and GPT-5-mini uses 30 in Chameleon and Undercover.}
\label{tab:balance_selfplay}
\end{table}

\paragraph{Result.}
The side that wins more often in self-play depends on the model.
As werewolves, GPT-5-mini and GPT-4o win 37\% and 40\% of games, while every other model wins 50 to 90\%.
Resistance likewise favors the majority for some models and the minority for others.
The two evaluation settings can differ substantially: GPT-5-mini's werewolves win 89\% against other tournament models but 37\% in self-play.
Because the two settings can differ this much, \Cref{sec:experiments} evaluates prompting methods in self-play only, against the same model without added text.

\section{Compute and API Cost}
\label{sec:appendix:cost}

At \Cref{tab:cost} we provide approximate API expenditures so that future work can budget similar experiments.
Costs are computed from the logged token usage at list prices, without prompt caching; for early runs that did not log usage, we use the mean per-turn cost of the same game and model.

\begin{table}[h]
\centering
\small
    \begin{tabular}{lr}
    \toprule
    \textbf{Component} & \textbf{Cost (USD)} \\
    \midrule
    Leaderboard tournament & $\sim$\$170 \\
    \textsc{SPaRTan} experiments (\Cref{sec:experiments}) & $\sim$\$300 \\
    Self-play baselines & $\sim$\$50 \\
    Discussion experiment & $\sim$\$40 \\
    Random-player games & $\sim$\$50 \\
    Outside benchmarks & $\sim$\$20 \\
    Playbook writing and alliance labels & $<$\$10 \\
    Open-weight models and judge & local GPU only \\
    \midrule
    \textbf{Total spend} & \textbf{$\sim$\$630} \\
    \bottomrule
    \end{tabular}

\caption{Approximate API cost by experiment phase.
The open-weights component runs on local GPUs and is excluded from the dollar total.
Costs cover the experiments reported in this paper.}
\label{tab:cost}
\end{table}

\end{document}